\documentclass[10pt,twocolumn,letterpaper]{article}

\usepackage[pagenumbers]{cvpr} % To force page numbers, e.g. for an arXiv version

\usepackage{graphicx}
\usepackage{amsmath}
\usepackage{amssymb}
\usepackage{booktabs}
\usepackage[acronym,xindy,toc]{glossaries}
\usepackage{bm} % bold math symbols
\usepackage{tabularx} % tables
\usepackage{booktabs} % booktabs tables
\usepackage{makecell} % formatting table cells
\usepackage{mathtools}
\usepackage{multirow}
\usepackage[pagebackref,breaklinks,colorlinks]{hyperref}

\renewcommand{\vec}[1]{\text{\textbf{#1}}}
\newcommand{\mat}[1]{\text{\textbf{#1}}}
\newcommand{\interval}[1]{{\footnotesize\textpm#1}}
\newcommand{\pred}[1]{{#1}^\mathrm{p}}
\newcommand{\gt}[1]{{#1}^\mathrm{gt}}
\newcommand{\Mpred}{\pred{\mathcal{M}}}
\newcommand{\Mgt}{\gt{\mathcal{M}}}
\newcommand{\Ppred}{\pred{\mathcal{P}}}
\newcommand{\Pgt}{\gt{\mathcal{P}}}
\newcommand{\stimes}{{\times}} % No space between axb

\DeclareMathOperator*{\argmin}{arg\,min}
\DeclarePairedDelimiter\abs{\lvert}{\rvert}%
\DeclarePairedDelimiter\norm{\lVert}{\rVert}%

\usepackage{comment}
\usepackage[capitalize]{cleveref}
\crefname{subsection}{Sec.}{Secs.}
\Crefname{subsection}{subsection}{Sections}
\Crefname{table}{Table}{Tables}
\crefname{table}{Tab.}{Tabs.}

\newcommand{\rough}[1]{{\leavevmode\color[rgb]{0.,0.7,0.7}#1}} % rough preliminary formulations/suggestions, feel free to improve them
\newcommand\blfootnote[1]{%
  \begingroup
  \renewcommand\thefootnote{}\footnote{#1}%
  \addtocounter{footnote}{-1}%
  \endgroup
}
\newacronym[shortplural={D$_{\text{dye}}$}, longplural={donor dye, ex. Alexa 488}]{ddye}{D$_{\text{dye}}$}{donor dye, ex. Alexa 488}

\newacronym[description={\glslink{r0}{F\"{o}rster distance}}]{R0}{$R_{0}$}{F\"{o}rster distance}

\newglossaryentry{r0}
{
  name=\glslink{R0}{\ensuremath{R_{0}}},
  text=F\"{o}rster distance,
  description={F\"{o}rster distance, where 50\% ...}, 
  sort=R
}

\newglossaryentry{kdeac}
{
  name=\glslink{R0}{\ensuremath{k_{DEAC}}},
  text=$k_{DEAC}$, 
  description={is the rate of deactivation from ... and emission)}, 
  sort=k
}

\newacronym[shortplural={TUM}, longplural={Technical University of Munich}]{tuma}{TUM}
{Technical University of Munich}

\newglossaryentry{tum}
{
  name = TUM,
  description = {A university in Germany, Bavaria, in the city of Munich},
  plural = TUM
}

\newglossaryentry{computer}
{
  name=computer,
  description={is a programmable machine that receives input,
               stores and manipulates data, and provides
               output in a useful format}
}

\newglossaryentry{nonlin}
{
	name={\glslink{nonlin}\ensuremath{h}},
	description={A nonlinearity}
}

\newacronym[shortplural={NNs}, longplural={neural networks}]{nn}{NN}{neural network}
\newacronym[shortplural={GNNs}, longplural={graph neural networks}]{gnn}{GNN}{graph neural network}
\newacronym[shortplural={CNNs}, longplural={convolutional neural networks}]{cnn}{CNN}{convolutional neural network}
\newacronym{ai}{AI}{artificial intelligence}
\newacronym{dl}{DL}{deep learning}
\newacronym{iou}{IoU}{Intersection over Union}
\newacronym{asp}{ASP}{Anatomic Segmentation using Proximity}
\newacronym{clasp}{CLASP}{Contrained Laplacian-based ASP}
\newacronym{csf}{CSF}{cerebrospinal fluid}
\newacronym{mri}{MRI}{magnetic resonance imaging}
\newacronym{pve}{PVE}{partial volume effect}
\newacronym{sdf}{SDF}{signed distance function}
\newacronym{lns}{LNS}{learned neighborhood sampling}
\newacronym{ct}{CT}{computer tomography}
\newacronym{hd}{HD}{90-percentile Hausdorff distance}
\newacronym{assd}{ASSD}{average symmetric surface distance}
\newacronym{sgd}{SGD}{stochastic gradient descent}
\newacronym{relu}{ReLU}{rectified linear unit}
\newacronym{exprecs}{EXPRECS}{\textbf{EXP}licit \textbf{RE}construction of \textbf{C}ortical \textbf{S}urfaces}
\newacronym{trt}{TRT}{test-retest dataset}
\newacronym{icp}{ICP}{iterative closest-point algorithm}
\newacronym{fcnn}{F-CNN}{fully-convolutional neural network}
\newacronym{mlp}{MLP}{multi-layer perceptron}
\newacronym{mr}{MR}{magnetic resonance}

\newglossaryentry{X}{name={\ensuremath{\mathcal{X}}},description={Input domain of a neural network.}}
\newglossaryentry{Y}{name={\ensuremath{\mathcal{Y}}},description={Output domain of a neural network.}}
\newglossaryentry{L}{name={\ensuremath{\mathcal{L}}},description={Loss function.}}
\newglossaryentry{theta}{name={\ensuremath{\theta}},description={Network parameters.}}
\newglossaryentry{dCD}{name={\ensuremath{d_\mathrm{CD}}},description={Chamfer distance}}
\newglossaryentry{LCD}{name={\ensuremath{\mathcal{L}_\mathrm{CD}}},description={Chamfer loss}}
\newglossaryentry{LCDcurv}{name={\ensuremath{\mathcal{L}_{\mathrm{C}}}},description={Chamfer loss}}
\newglossaryentry{Jacc}{name={\ensuremath{J}},description={Jaccard index}}
\newglossaryentry{dJacc}{name={\ensuremath{d_{\mathrm{Jacc}}}},description={Jaccard distance}}
\newglossaryentry{Dice}{name={\ensuremath{DSC}},description={Dice coefficient}}
\newglossaryentry{Hausd}{name={\ensuremath{d_\mathrm{H}}},description={Hausdorff distance}}
\newglossaryentry{dASSD}{name={\ensuremath{d_\mathrm{AD}}},description={Average symmetric surface distance}}
\newglossaryentry{LCE}{name={\ensuremath{\mathcal{L}_\mathrm{BCE}}},description={Cross entropy loss}}
\newglossaryentry{LNC}{name={\ensuremath{\mathcal{L}_{\mathrm{n}, \, intra}}},description={Normal consistency loss}}
\newglossaryentry{Lcos}{name={\ensuremath{\mathcal{L}_{\mathrm{n}, \, inter}}},description={Cosine loss}}
\newglossaryentry{LLapabs}{name={\ensuremath{\mathcal{L}_{\mathrm{Lap}, \, abs}}},description={Laplacian smoothing loss w.r.t. absolute vertex coordinates}}
\newglossaryentry{LLaprel}{name={\ensuremath{\mathcal{L}_{\mathrm{Lap}, \, rel}}},description={Laplacian smoothing loss w.r.t. relative vertex coordindates}}
\newglossaryentry{LLap}{name={\ensuremath{\mathcal{L}_{\mathrm{Lap}}}},description={Laplacian smoothing loss}}
\newglossaryentry{Ledge}{name={\ensuremath{\mathcal{L}_\mathrm{edge}}},description={Edge loss}}
\newglossaryentry{phi}{name={\ensuremath{\bm{\phi}}},description={Feature vector}}
\newglossaryentry{diff_co}{name={\ensuremath{\bm{\xi}}},description={Differential coordinates}}
\newglossaryentry{curv}{name={\ensuremath{\bar{\kappa}}},description={Discrete mean curvature}}
\newglossaryentry{Diff_co}{name={\ensuremath{\bm{\Xi}}},description={Matrix of differential coordinates}}
\newglossaryentry{Lap}{name={\ensuremath{\bm{\mathrm{L}}}},description={Laplace operator}}
\newglossaryentry{Lap-u}{name={\ensuremath{\bm{\mathrm{L}}_\mathrm{u}}},description={Laplace operator with uniform weights}}
\newglossaryentry{Lap-cot}{name={\ensuremath{\bm{\mathrm{L}}_\mathrm{cot}}},description={Laplace operator with cotangent weights}}
\newglossaryentry{A}{name={\ensuremath{\mathrm{\textbf{A}}}},description={Adjacency matrix}}
\newglossaryentry{T}{name={\ensuremath{\mathcal{T}}},description={Template mesh}}
\newglossaryentry{D}{name={\ensuremath{\mathrm{\textbf{D}}}},description={Degree matrix}}
\newglossaryentry{n}{name={\ensuremath{n}},description={Number of mesh vertices}}
\newglossaryentry{deg}{name={\ensuremath{\mathrm{deg}}},description={Degree of a vertex}}
\newglossaryentry{N}{name={\ensuremath{\mathcal{N}}},description={Neighborhood of a vertex}}
\newglossaryentry{N_img}{name={\ensuremath{N}},description={The number of pixels or voxels in an image}}
\newglossaryentry{H_img}{name={\ensuremath{H}},description={The height of an image}}
\newglossaryentry{W_img}{name={\ensuremath{W}},description={The width of an image}}
\newglossaryentry{D_img}{name={\ensuremath{D}},description={The depth of an image}}
\newglossaryentry{V}{name={\ensuremath{\mathcal{V}}},description={Set of vertices}}
\newglossaryentry{Vmat}{name={\ensuremath{\bm{\mathrm{V}}}},description={Vertices stacked into a matrix}}
\newglossaryentry{Fmat}{name={\ensuremath{\bm{\mathrm{F}}}},description={Faces stacked into a matrix}}
\newglossaryentry{E}{name={\ensuremath{\mathcal{E}}},description={Set of edges}}
\newglossaryentry{F}{name={\ensuremath{\mathcal{F}}},description={Set of faces}}
\newglossaryentry{lr}{name={\ensuremath{\lambda}},description={Learning rate}}
\newglossaryentry{Lvox}{name={\ensuremath{\mathcal{L}_\mathrm{vox}}},description={Voxel-based loss}}
\newglossaryentry{Lmesh}{name={\ensuremath{\mathcal{L}_\mathrm{mesh}}},description={Mesh-based loss}}
\newglossaryentry{Lmesh_cons}{name={\ensuremath{\mathcal{L}_{\mathrm{mesh},\, cons}}},description={Geometry-consistency loss}}
\newglossaryentry{Lmesh_reg}{name={\ensuremath{\mathcal{L}_{\mathrm{mesh}, \, reg}}},description={Mesh-regularization loss}}
\newglossaryentry{S}{name={\ensuremath{S}},description={The number of mesh-deformation stages}}
\newglossaryentry{n_seg}{name={\ensuremath{L}},description={The number of segmentation outputs (voxel-based)}}
\newglossaryentry{n_struc}{name={\ensuremath{C}},description={The number of surfaces or connected components in one mesh}}
\newglossaryentry{Vrel}{name={\ensuremath{\bm{\Delta}}},description={Matrix of relative coordinates}}
\newglossaryentry{mlw}{name={\ensuremath{\lambda}},description={Mesh-loss weight}}

\def\cvprPaperID{7672} % *** Enter the CVPR Paper ID here
\def\confName{CVPR}
\def\confYear{2022}

\begin{document}

%%%%%%%%% TITLE

\title{Vox2Cortex: Fast Explicit Reconstruction of Cortical Surfaces from \\ 3D MRI Scans with Geometric Deep Neural Networks}

\author{Fabian Bongratz$^{1\dagger*}$,
Anne-Marie Rickmann$^{1,2*}$,
Sebastian P{\"{o}}lsterl$^{2}$,
Christian Wachinger$^{1,2}$
\\$^{1}$Technical University of Munich,
$^{2}$Ludwig-Maximilians-University Munich
\\Lab for Artificial Intelligence in Medical Imaging
}

\maketitle
\blfootnote{$*$: Equal contribution, $\dagger$: Corr.~author (fabi.bongratz@tum.de)}

%%%%%%%%% ABSTRACT
\begin{abstract}
   The reconstruction of cortical surfaces from brain magnetic resonance imaging (MRI) scans is essential for quantitative analyses of cortical thickness and sulcal morphology. 
   Although traditional and deep learning-based algorithmic pipelines exist for this purpose, they have two major drawbacks: lengthy runtimes of multiple hours (traditional) or intricate post-processing, such as mesh extraction and topology correction (deep learning-based). In this work, we address both of these issues and propose Vox2Cortex, a deep learning-based algorithm that directly yields topologically correct, three-dimensional meshes of the boundaries of the cortex. Vox2Cortex leverages convolutional and graph convolutional neural networks to deform an initial template to the densely folded geometry of the cortex represented by an input MRI scan. We show in extensive experiments on three brain MRI datasets that our meshes are as accurate as the ones reconstructed by state-of-the-art methods in the field, without the need for time- and resource-intensive post-processing. To accurately reconstruct the tightly folded cortex, we work with meshes containing about 168,000 vertices at test time, scaling deep explicit reconstruction methods to a new level.
  
\end{abstract}

%%%%%%%%% BODY TEXT
\section{Introduction}
\label{sec:intro}

% Voxel-based segmentation of cortex popular and fast (recent advances in DL compared to traditional frameworks)

% Great advances in CV regarding mesh-based reconstruction algorithms

% Our contribution: Mesh-based cortex reconstruction/segmentation --> higher resolution than voxel-based methods
Over the last years, \glspl{fcnn} have been introduced for the automatic whole-brain segmentation of MRI scans. Architecture choices range from multi-view 2D F-CNNs~\cite{henschel2020,roy2019} and patch-based 3D F-CNNs~\cite{HUO2019105} to full-size 3D F-CNNs~\cite{rickmann2020recalibrating}. %cite sth. 
In contrast to time-consuming traditional methods for brain analysis, such networks perform the segmentation in seconds.
The resulting voxel-based segmentation can be adequately used for volume measurements of the whole brain or subcortical structures.
However, the cerebral cortex, a thin and highly-folded sheet of neural tissue of the brain, is not well captured by voxel-based segmentations since it is just a few voxels thick. 
Instead, cortical surfaces have been reconstructed as triangular meshes~\cite{fischl2012freesurfer} to capture the intrinsic two-dimensional structure, which enables accurate measurements like thickness, volume, and gyrification. 
Such \textit{in-vivo} measurements have not only been important for studying higher-level processes like consciousness, thought, emotion, reasoning, language, and memory~\cite{phillips1984localization,shipp2007structure}, but also for detecting cortical atrophy in numerous brain disorders like Alzheimer's disease~\cite{lerch2005} and schizophrenia~\cite{kuperberg2003}.

\begin{figure}[t]
    \centering
    \includegraphics[width=\linewidth]{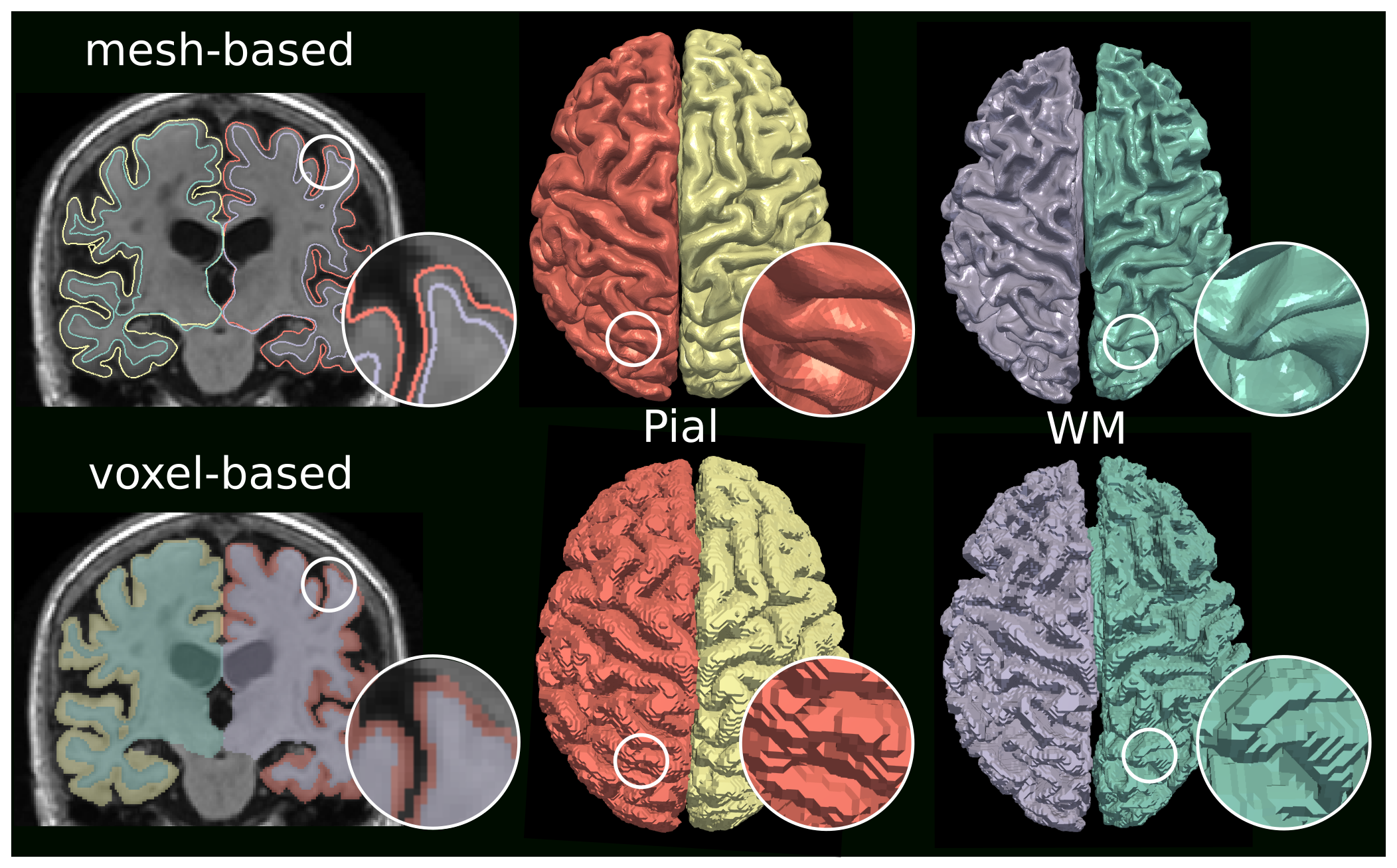}
    \caption{Coronal slices of a brain MRI scan with overlays of mesh- and voxel-based segmentation, and 3D renderings of corresponding pial and white matter (WM) surfaces. The top row shows meshes generated by our method, the bottom row shows meshes generated by applying marching cubes on the voxel segmentation. Marching cubes meshes clearly show stair-case artifacts.\vspace{-3mm}}
    \label{fig:overview}
\end{figure}

The current standard for cortical surface reconstruction is FreeSufer~\cite{dale1999,fischl2012freesurfer}, which produces smooth, accurate, and topologically correct surface meshes but runs for several hours per scan.  
On the other hand, voxel-based segmentation algorithms are quick and their outputs can be used to obtain mesh-based surface representations using algorithms like marching cubes~\cite{lewiner2003efficient,lorensen1987}. 
However, the extracted meshes suffer from typical stair-case artifacts and their resolution is limited by the resolution of the segmentation, respectively the resolution of the initial MRI scan, see Figure~\ref{fig:overview}. Further, the predicted segmentation might have topological defects and a topology-correction algorithm needs to be applied before or after marching cubes mesh extraction, see Figure~\ref{fig:compact_comparison}. This need for post-processing steps is not only cumbersome; while topology correction algorithms generate meshes with the desired topology, they do not always produce anatomically correct meshes~\cite{fischl2012freesurfer}, see also supplementary Figure~4.

Recently, seminal research in the field of single-view or multi-view reconstruction~\cite{wang2020pixel2mesh,wen2019}
introduced network architectures that can produce explicit surface representations from 2D images by transforming a template mesh (e.g., a sphere) using graph convolutions. 
These methods are promising for our task, as they can enforce the desired topology of the template mesh and directly output a mesh without the need for post-processing. So far, these methods have not been applied to reconstruct shapes with such complex folding patterns as the cerebral cortex.
%complex shapes like cortical surface meshes.

In this work, we introduce Vox2Cortex, a deep learning-based algorithm that allows for \emph{direct} reconstruction of \emph{explicit} meshes of cortical surfaces from brain MRI. 
Vox2Cortex  simultaneously predicts white matter (WM) and pial surfaces of both hemispheres, resulting in an output of four meshes. 
Vox2Cortex enforces a spherical topology to prevent the formation of holes or handles in the mesh and does therefore not need post-processing like~\cite{santacruz2021}.
We use up to 168,000 vertices per mesh to capture the highly complex folding patterns of the cortex with high accuracy. 
This number is substantially higher than in prior work on single-view reconstruction that typically uses less than 10,000 vertices per surface.
Our main contributions are:
\begin{itemize}
    \item Vox2Cortex is the first network specifically designed to extract explicit surface representations of the cortex from MRI scans using a combination of a convolutional neural network and a graph neural network.
    \item We ensure spherical topology of the generated meshes by learning to deform a template mesh with fixed topology and arbitrary resolution.
    \item Vox2Cortex models the interdependency between WM and pial surfaces by exchanging information between them.
    \item We propose a new loss function for training explicit reconstruction methods that leverages local curvature to weight Chamfer distances. It turns out that this loss is crucial for learning to reconstruct the tightly folded surfaces of the human cortex.
    \item We demonstrate on multiple brain datasets that Vox2Cortex performs on par with or better than existing implicit and voxel-based reconstruction methods in terms of accuracy and consistency while being 25 times faster at inference time.
  
\end{itemize}
The main objective of our work is to develop a fast and highly accurate reconstruction of the cortex, which has high clinical value. Nevertheless, our contributions to predicting multiple, tightly folded meshes are generic and can also be applied to other applications. Our code is publicly available at \url{https://github.com/ai-med/Vox2Cortex}.

\begin{comment}
In more detail, our main contributions are:
\begin{itemize}
    \item We combine multiple advances in the field of deep explicit surface reconstruction in one end-to-end trainable architecture. Moreover, we demonstrate on multiple brain datasets that their combination leads to an algorithm that performs \rough{on par} with existing implicit and voxel-based reconstruction methods in terms of accuracy and consistency while being \rough{considerably} faster at inference time.
    \item We provide an extensive ablation study of the used architecture, uncovering the individual contribution of the different architectural building blocks. This is, in particular, necessary since previous work often lacks an analysis of architectural choices, e.g., sampling either from the encoder or decoder without elaborating \emph{why} one is preferred over another.
    \item We propose a new loss function for training explicit reconstruction methods that leverages local curvature to weight Chamfer distances. It turns out that this loss is crucial for learning to reconstruct the tightly folded surfaces of the human cortex.
    \item We demonstrate the practical applicability of our method on downstream tasks, namely measurement of cortical thickness and surface area.
\end{itemize}

\end{comment}

\section{Related Work}
\begin{figure}[t]
    \centering
    \includegraphics[width=\linewidth]{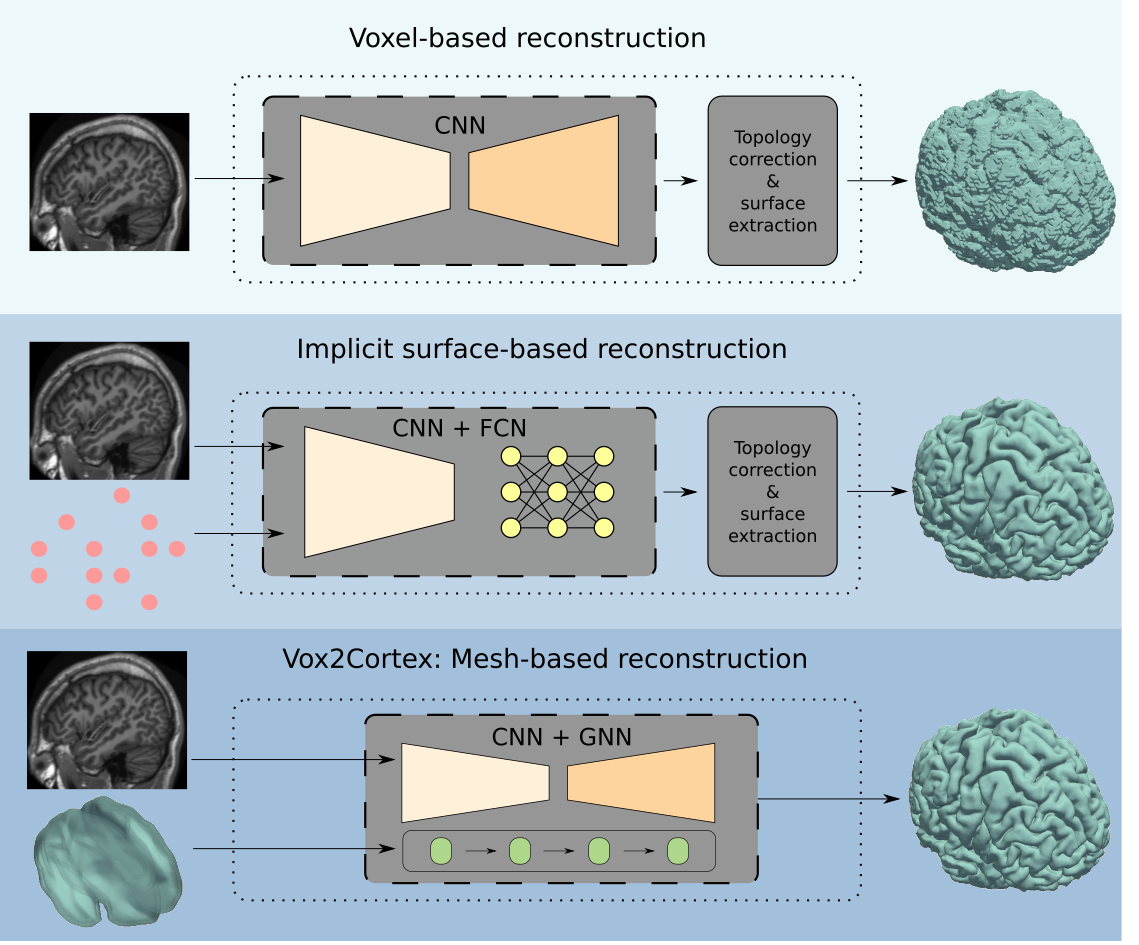}
    \caption{Existing deep learning-based approaches for cortical surface reconstruction from MRI usually rely on an implicit or voxel representation of the cortex. Those methods require intricate post-processing steps, including topological correction and marching cubes~\cite{lewiner2003efficient,lorensen1987} to generate surface meshes. These meshes are, however, crucial for downstream applications such as the measurement of cortical thickness. In contrast, our model directly yields highly accurate meshes of the WM and the pial surfaces.\vspace{-3mm}}
    \label{fig:compact_comparison}
\end{figure}
Traditional brain MRI processing pipelines~\cite{dale1999cortical,fischl1999cortical,smith2004advances} consist of many steps including registration, segmentation, and cortical surface extraction. 
These pipelines are computationally intensive and, therefore, delay the availability of cortex measures after scan acquisition. %expensive and, therefore, not suitable for clinical use.
Current deep learning approaches for cortical surface reconstruction focus on voxel-based or implicit surface reconstruction methods. Instead, we propose a mesh-based approach. We depict an overview of the different approaches in Figure~\ref{fig:compact_comparison} and review related work in the following. 

\textbf{Voxel-based surface reconstruction}:
FastSurfer~\cite{henschel2020} is a deep learning approach that speeds up the FreeSurfer pipeline but focuses on voxel-based segmentation and requires the use of marching cubes for surface generation and topology correction.
SegRecon~\cite{gopinath2021} proposes a 3D CNN for joint segmentation and surface reconstruction, which learns a 3D \gls{sdf} and therefore still requires marching cubes and topology correction.

\textbf{Deep implicit representations}:
Deep implicit representations have become a popular research field in 3D computer vision in recent years~\cite{mescheder2019,park2019deepsdf,sitzmann2020implicit,xu2019}. The idea behind this is to learn a function that maps 3D coordinates to a continuous implicit representation of shape, usually an \gls{sdf}. 
DeepCSR~\cite{santacruz2021} was explicitly proposed for cortical surface reconstruction.  
As their shape representation is continuous, meshes of any desired resolution can be obtained. DeepCSR achieves state-of-the-art results in terms of distance metrics but still requires marching cubes and topology correction.

% surface reconstruction:
\textbf{Mesh-based surface reconstruction}:
Mesh-deformation networks~\cite{kong2021b,kong2021,wang2020pixel2mesh,wen2019,wickramasinghe2020} are learning deformable mesh models that take as input a template mesh or sphere initialization and the image and iteratively deform the mesh by learning the deformation field of the vertices. 
Voxel2Mesh~\cite{wickramasinghe2020} and MeshDeformNet~\cite{kong2021b,kong2021} 
were applied to medical data and showed promising results, albeit mainly applied to relatively simple shapes like the hippocampus, liver, or heart. As the cortex is a significantly more complicated shape, it is still to be evaluated if this approach can work well on cortex reconstruction.
PialNN~\cite{ma2021b} learns to deform an initial WM surface to a pial surface by a series of graph convolutions but requires FreeSurfer to create the WM surface.

\section{Method}
\label{sec:method}

%\FB{[I thought about calling the method \gls{exprecs}, but alternative propositions are welcome as well :)]}

% !TeX root = ../main.tex
% !TeX spellcheck = en_US

In this section, we provide a detailed description of Vox2Cortex, % \emph{\acrshort{exprecs}} (\acrlong{exprecs}), 
a method for fast cortical surface reconstruction from 3D \gls{mr} images based on geometric deep learning.

%\subsection{\gls{exprecs} Architecture}
\subsection{Vox2Cortex Architecture}
\begin{figure*}[t]
    \centering
    \includegraphics[width=\textwidth]{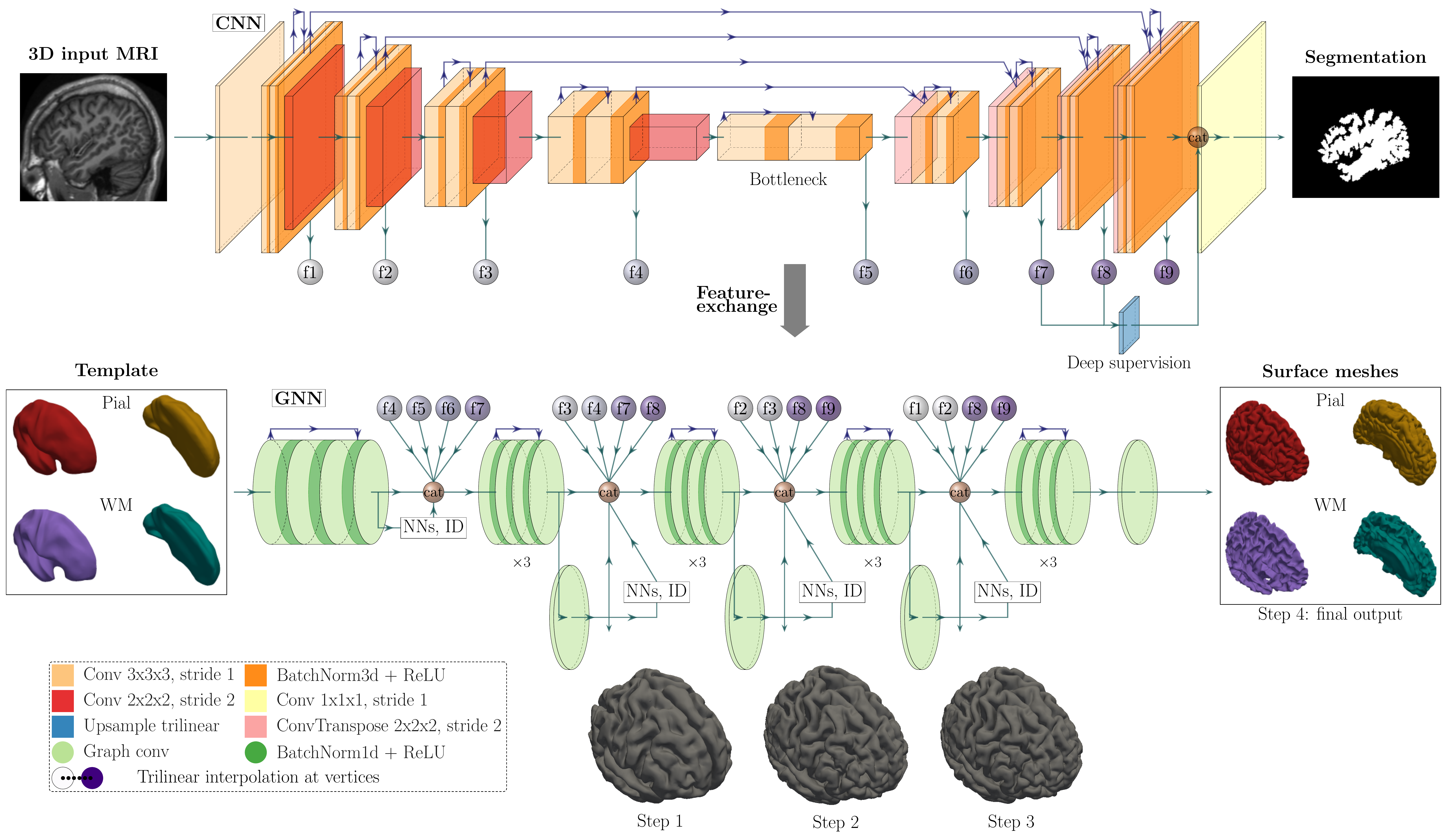}
    \caption{Vox2Cortex pipeline. The architecture takes as input a 3D brain scan and a mesh template and predicts a voxel-wise segmentation map and cortical surface meshes. The principal building blocks are a \gls{cnn} and a \gls{gnn}. The \gls{gnn} deforms the initial template within four deformation steps based on image- and shape-descriptive features to the predicted output meshes. We created the drawing using~\cite{iqbal2020}.\vspace{-3mm}}
	\label{fig:exprecs}
\end{figure*}

%Our \gls{exprecs} network 
Vox2Cortex takes a 3D brain MRI scan and a mesh template as input and computes simultaneously four cortical surfaces, namely the white-matter and the pial surfaces of each hemisphere. As a side product, the network predicts a voxel-wise brain segmentation (i.e., of the gray matter and the tissue enclosed by the gray matter). 
Inspired by previous related methods~\cite{kong2021b,kong2021,wang2018, wickramasinghe2020}, our architecture consists of two neural sub-networks, a \gls{cnn} that operates on voxels and a \gls{gnn} responsible for mesh deformation. Both networks are connected via feature-sampling modules that map features extracted by the \gls{cnn} to vertex locations of the meshes. 
\Cref{fig:exprecs}  depicts the whole architecture together with exemplary in- and outputs. 
In the following, we detail the individual building blocks: image encoder, image decoder,  mesh deformation network, and information exchange.

\paragraph{Image encoding and decoding (\gls{cnn})}

We use a residual UNet architecture~\cite{cicek2016,isensee2020,ronneberger2015,zhang2018} for image-feature extraction. This encoder-decoder \gls{fcnn} solely takes the 3D brain scan as input and segments it into brain and background voxels.
The UNet consist of multiple residual convolution blocks with batch-normalization layers~\cite{ioffe2015} and ReLU activations~\cite{nair2010}. For up- and downsampling of the feature maps, we use (transposed) convolutional layers with stride 2. Moreover, we add deep-supervision branches~\cite{zeng2017} to propagate the segmentation loss directly to lower decoder layers. 
\Cref{fig:exprecs} illustrates the network in detail. %For more details about this part of the network, please refer to .

\paragraph{Mesh deformation (\gls{gnn})}
The second major building block of our architecture is a \gls{gnn} that takes a template mesh as input~\cite{kong2021b,kong2021,wang2018, wickramasinghe2020}. 
% Sentence added after rebuttal
Its precedence over an MLP~\cite{wang2019, groueix2018} is validated by our ablation study (cf.~\Cref{tab:ablation study}).
The \gls{gnn} deforms the template in four mesh-deformation steps that build upon each other, see \Cref{fig:exprecs}. Each deformation step predicts vertex-specific displacement vectors relative to the mesh produced by the previous step.
Similar to the image-processing \gls{cnn}, our \gls{gnn} sub-architecture consists of multiple residual blocks, allowing for effective residuum-based learning~\cite{he2016, kong2021}. Each graph-residual block consists of three graph convolutions with subsequent batch-norm and ReLU operations. The input residuum is added before the last ReLU operation and potentially reshaped with nearest-neighbor interpolation. Following~\cite{kong2021}, the initial graph-residual block has a large number of channels compared to later blocks to weight the extracted vertex features similar to the image-based features in terms of feature-vector length. Ultimately, a single graph convolutional layer is used to output the displacement vectors for each vertex at the current deformation stage. 

% Graph convolutions
We leverage spectrum-free graph convolutions, in the flavor of message-passing operations~\cite{bronstein2021}, and choose the implementation provided in~\cite{ravi2020}.
%among the many variants of graph convolutions, e.g.,~\cite{defferrard2016, hamilton2017,kipf2017,morris2019,wu2019}. 
Formally, each graph convolutional layer transforms the features from a previous layer $\vec{f}_i \in \mathbb{R}^{d_\mathrm{in}}$ of a vertex $\vec{v}_i \in \gls{V}$ by aggregating
\begin{equation} \label{eq:graph conv}
	\vec{f}'_i =  \frac{1}{1 + \abs{\gls{N}(i)}}\left[\mat{W}_0 \vec{f}_i + \vec{b}_0 + \sum_{j \in \gls{N}(i)} (\mat{W}_1 \vec{f}_j + \vec{b}_1)\right],
\end{equation}
where $\mat{W}_0, \mat{W}_1 \in \mathbb{R}^{d_\mathrm{out} \stimes d_\mathrm{in}}$ together with $\vec{b}_0, \vec{b}_1 \in \mathbb{R}^{d_\mathrm{out}}$ represent linear transformations and $\gls{N}(i)$ is the set of neighbors of $\vec{v}_i$. Besides the mesh-output layers, graph convolutional layers are followed by batch-norm and ReLU layers in our network.

\paragraph{Image features} While \glspl{cnn} and \glspl{gnn} are established network architectures, so far there have only been few attempts for combining them~\cite{wang2018, wickramasinghe2020, kong2021, kong2021b} and the question regarding the best way to exchange information between them has not been answered yet. We combine feature maps from the UNet encoder as well as the UNet decoder at multiple resolutions.
This allows the network to identify relevant information during the training process, which is not possible with a narrow pre-selection of feature maps. In order to get features in continuous 3D space, we interpolate the feature maps from discrete voxels trilinearly. 

\vspace{-1.5mm}
\paragraph{Inter-mesh neighbors}
To the best of our knowledge, deep explicit reconstruction methods have been only applied for the reconstruction of \emph{isolated} objects so far. However, we deal with multiple surfaces, among which we have an \emph{interdependence} as the inner and outer brain surfaces are always aligned. 
To improve the quality of the reconstruction, we model the interdependence of the meshes by exchanging information between them. 
%Therefore, we propose to exchange information, in the form of coordinates of the closest neighbors from the respective other mesh, between them during training. 
To this end, we concatenate the coordinates of the five nearest outer vertices and a surface identifier to the features of each inner vertex (and vice versa) before every mesh-deformation block. Traditional methods like FreeSurfer~\cite{dale1999cortical} also model this inter-dependency by extracting an outer surface based on the inner one (and not simultaneously as we do).

\paragraph{Mesh template} \label{sec:template}
Existing explicit reconstruction methods usually start from simple mesh templates like spheres or ellipsoids. 
To improve the cortex reconstruction quality, we propose using a more application-specific template. 
We created four genus-0 templates by choosing a random FreeSurfer mesh %from the MALC dataset~\cite{landman2012} 
and applying Laplacian smoothing~\cite{vollmer1999} until the surface did not change anymore. 
\Cref{fig:exprecs} illustrates the template as input to the network. 

%Interestingly, we found that it is possible to use a smaller template during training compared to testing. 
Importantly, we observed that it is possible to use a template with fewer vertices during training compared to testing and thereby increase the surface accuracy of a trained model.
In other words, we can choose the desired mesh resolution independently of the template used during training. In our experiments, we used about 42,000 vertices per surface during training, whereas we increase this number to about 168,000(!) during testing. Across all four surfaces, this yields a total over 672,000 vertices.

\subsection{Loss Functions}
In this section, we describe the loss function for training the Vox2Cortex network with particular emphasis on our novel \emph{curvature-weighted Chamfer loss}. Definitions of the other loss functions are in the supplementary material.

Let $\pred{y} = \{\{\Mpred_s, \pred{\gls{Vrel}}_s, \pred{B}_l\} | s = 1, \ldots, \gls{S}; \, l = 1, \ldots, \gls{n_seg}\}$ be the prediction of our model, where $\Mpred_s = (\pred{\gls{V}}_s, \pred{\gls{F}}_s, \pred{\gls{E}}_s)$ are the predicted meshes at \gls{S} different deformation stages, $\pred{\gls{Vrel}}_s$ are the predicted displacement vectors stacked into a tensor, and $\pred{B}_l \in [0, 1]^{\gls{H_img}  \gls{W_img}  \gls{D_img}}$ are the voxel-wise binary-segmentation maps (final prediction of the \gls{cnn} and $\gls{n_seg}-1$ deep-supervision outputs). In general, each mesh $\Mpred_s$ consists again of \gls{n_struc} different surfaces, i.e., $\Mpred_s = \{\Mpred_{s,c} | c = 1, \ldots, \gls{n_struc}\}$. In our case, $\gls{n_struc}$ is equal to four (WM and pial surfaces for each hemisphere). As we train our model in a supervised manner, we assume having corresponding ground-truth meshes and segmentation maps $\gt{y} = (\Mgt, \gt{B})$ available for each sample in the training set. 

\paragraph{Total loss}
The loss function of Vox2Cortex  consists of a voxel $\gls{Lvox}$ and a mesh \gls{Lmesh} part
\begin{equation}
	\gls{L}(\pred{y}, \gt{y}) = \gls{Lvox}(\pred{y}, \gt{y}) + \gls{Lmesh}(\pred{y}, \gt{y}).
\end{equation}

\paragraph{Voxel loss}
Let $\gls{LCE}(\pred{B}_l, \gt{B})$ be the binary cross-entropy loss between a predicted segmentation map $\pred{B}_l \in [0, 1]^{\gls{H_img}  \gls{W_img}  \gls{D_img}}$ and a label $\gt{B} \in \{0, 1\}^{\gls{H_img}  \gls{W_img}  \gls{D_img}}$. Taking into account that we have \gls{n_seg} segmentation outputs (the final segmentation of the \gls{cnn} as well as $\gls{n_seg}-1$ deep-supervision outputs), we compute the voxel loss as
\begin{equation} \label{eq:voxel loss}
	\gls{Lvox}(\pred{y}, \gt{y}) = \sum_{l=1}^{\gls{n_seg}} \gls{LCE}(\pred{B}_l, \gt{B}).
\end{equation}

%\paragraph{Mesh-Based Loss}
\paragraph{Mesh loss}
Compared to the voxel loss, it is more challenging to define an adequate loss function between two meshes. 
This is partly due to the lack of point correspondences between the predicted meshes and the labels, making it complicated to define a loss function that enforces a ``good'' cortical surface mesh.
Inspired by previous explicit reconstruction methods~\cite{kong2021,wang2018, wickramasinghe2020}, we use a combination of geometry-consistency and regularization losses
\begin{equation} \label{eq:mesh loss}
		\gls{Lmesh}(\pred{y}, \gt{y}) = \gls{Lmesh_cons}(\pred{y}, \gt{y}) + \gls{Lmesh_reg}(\pred{y}).
\end{equation} 
For the geometry-consistency loss, we add up a novel curvature-weighted Chamfer loss $\gls{LCDcurv}$, further described in the next section, and an inter-mesh normal consistency (also called normal distance~\cite{gkioxari2019}) $\gls{Lcos}$
\begin{align} \label{eq:cons loss}
\begin{split}
    	\gls{Lmesh_cons}(\pred{y}, \gt{y}) 
    	&= \sum_{s=1}^{S} \sum_{c=1}^{C} \left[ \gls{mlw}_{1,c} \, \gls{LCDcurv}(\Mpred_{s,c}, \Mgt_c) \right. \\
    	&+ \left. \gls{mlw}_{2,c} \, \gls{Lcos}(\Mpred_{s,c}, \Mgt_c) \right].
\end{split}
\end{align}
The regularization loss consists of Laplacian smoothing of the displacement fields $\gls{LLaprel}$, intra-mesh normal consistency $\gls{LNC}$,  and edge length $\gls{Ledge}$
\begin{align} \label{eq:reg loss}
\begin{split}
		\gls{Lmesh_reg}(\pred{y}) 
		&= \sum_{s=1}^{S} \sum_{c=1}^{C} \left[ \gls{mlw}_{3,c} \, \gls{LLaprel}(\Mpred_{s,c}, \pred{\gls{Vrel}}_{s,c}) \right. \\
		&+{} \gls{mlw}_{4,c} \, \gls{LNC}(\Mpred_{s,c}) \\
		&+{} \left. \gls{mlw}_{5,c} \, \gls{Ledge}(\Mpred_{s,c}) \right].
\end{split}
\end{align}
More details about the loss hyperparameters can be found in the supplementary material.

\paragraph{Curvature-weighted Chamfer loss}
\begin{figure}[t]
    \centering
    \includegraphics[width=\linewidth]{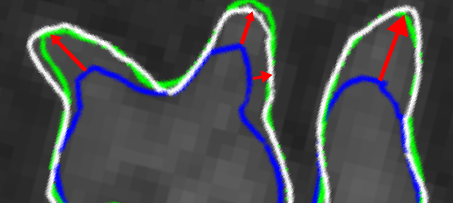}
    \caption{Our curvature-weighted Chamfer loss (green) often leads to a more accurate model of the cortical folds compared to the standard Chamfer distance (blue). FreeSurfer ground truth (white).\vspace{-3mm}}
    \label{fig:effect_curv}
\end{figure}

The Chamfer distance has become an essential building block for learning deformable-shape models of surfaces~\cite{fan2017, wang2018}, but needs to be combined with regularization terms, as it can otherwise lead to edge intersections~\cite{gupta2020}.
In regions with high curvature, as often found in the densely folded geometry of the cortex, the smoothing effect of the regularization terms can lead to lower geometric accuracy.
Hence, we propose a curvature-weighted Chamfer loss function that emphasizes high-curvature regions and improves the reconstruction of densely folded areas, cf.~\Cref{fig:effect_curv}. 
A proof showing that this loss ``pushes'' predicted points in high-curvature regions more towards their correct position compared to points in low-curvature regions, under mild assumptions, is in the supplementary material.  
Note that we only use curvature weights from ground-truth points as the potentially wrong curvature in a prediction could be misleading. 

Assume we have a curvature function $\kappa(\vec{p}) \in \mathbb{R}_{\geq 0}$ assigning a curvature to a point $\vec{p}$, then we define the curvature-weighted-Chamfer loss as
\begin{align} \label{eq:weighted_Chamfer distance}
	\begin{split}	
	\gls{LCDcurv} (\Mpred_{s,c}, \Mgt_c) 
	&= \frac{1}{\abs{\Pgt_c}} \sum_{\vec{u} \in \Pgt_c} \kappa(\vec{u}) \underset{\vec{v} \in \Ppred_{s,c}}{\min} \norm{\vec{u} - \vec{v}}^2 \\
	&+{} \frac{1}{\abs{\Ppred_{s,c}}} \sum_{\vec{v} \in \Ppred_{s,c}} \kappa(\tilde{\vec{u}}) \underset{\vec{u} \in \Pgt_c}{\min} \norm{\vec{v} - \vec{u}}^2,
	\end{split}
\end{align}
where $\tilde{\vec{u}} = \underset{\vec{r} \in \Pgt_c}{\argmin} \norm{\vec{v} - \vec{r}}^2$. In practice, we found that
\begin{equation}
	\kappa(\vec{p}) = \min\{1 + \gls{curv}(\vec{p}), \, \kappa_{\max}\}, 
\end{equation}
where $\gls{curv}(\vec{p})$ is the discrete mean curvature~\cite{meyer2003, nealen2006}, is a good choice ($\kappa_{\max} = 5$ in our experiments). During training, every $\Pgt_c|_{c=1, \ldots, C}$ contains as many vertices as the smallest ground-truth surface in the training set. On the other hand, we sample the point clouds $\Ppred_c|_{c=1, \ldots, C}$ in a differentiable manner from the surface of the predicted meshes~\cite{gkioxari2019,smith2019} such that they contain the same number of points as the reference meshes.

\section{Results}
\label{sec:experiments}
In the following, we compare Vox2Cortex to related approaches on multiple datasets. 
%and for several downstream tasks. 
In addition, we reveal the individual contributions of the main building blocks of Vox2Cortex in an extensive ablation study.

\subsection{Datasets} \label{sec:datasets}
\paragraph{Pre-processing}
All data used in this work was processed using FreeSurfer v5.3~\cite{fischl2012freesurfer}. We used the orig.mgz files and WM and pial surfaces generated by Freesurfer. 
We follow the pre-processing pipeline by~\cite{santacruz2021} for experiments with DeepCSR. We registered the MRI scans to the MNI152 space using rigid and subsequent affine registration. When used as supervised training labels, the FreeSurfer meshes were simplified to about 40,000 vertices per surface using quadric edge collapse decimation~\cite{garland1997}. We first padded all input images to have shape $192 \stimes 208 \stimes 192$ and then resized them to $128 \stimes 144 \stimes 128$ voxels. Intensity values were min-max-normalized to the range $[0, 1]$.

\begin{table*}
\footnotesize
\setlength{\tabcolsep}{4.6pt} % Reduce white-space between columns, SHOULD BE SET AFTER TABLE IS FINISHED

\begin{tabular}{lcccccccc}
    \toprule
    & \multicolumn{2}{c}{Left WM Surface} 
    & \multicolumn{2}{c}{Right WM Surface}
    & \multicolumn{2}{c}{Left Pial Surface}
    & \multicolumn{2}{c}{Right Pial Surface}
    \\
    \cmidrule(lr){2-3} \cmidrule(lr){4-5} \cmidrule(lr){6-7} \cmidrule(lr){8-9} 
    & ASSD  & HD & ASSD & HD  & ASSD & HD & ASSD & HD \\\midrule

     Vox2Cortex & \textbf{0.401} \interval{0.065} & \textbf{0.894} \interval{0.177} & \textbf{0.403} \interval{0.057} & \textbf{0.896} \interval{0.142} & \textbf{0.375} \interval{0.055} & 0.965 \interval{0.210} & \textbf{0.378} \interval{0.060} & 1.012 \interval{0.248} \\
    
    Vox2Cortex* & 0.455 \interval{0.063} & 1.057 \interval{0.195} & 0.457 \interval{0.056} & 1.055 \interval{0.145} & 0.467 \interval{0.057} & 1.316 \interval{0.278} & 0.470 \interval{0.0611} & 1.371 \interval{0.281} \\
    \midrule
    
    \addlinespace[0.3em]
    Voxel2Mesh* \cite{wickramasinghe2020} & 0.528 \interval{0.222} & 1.209 \interval{0.732} & 0.528 \interval{0.197} & 1.186 \interval{0.625} & 0.486 \interval{0.114} & 1.457 \interval{0.398} & 0.476 \interval{0.108} & 1.440 \interval{0.384} \\ 
    
    \addlinespace[0.3em]
    Encoder features & 0.453 \interval{0.072} & 0.984 \interval{0.177} & 0.456 \interval{0.054} & 1.007 \interval{0.144} & 0.432 \interval{0.067} & 1.057 \interval{0.211} & 0.430 \interval{0.059} & 1.040 \interval{0.174}\\ 
    
    \addlinespace[0.3em]
    Classic Chamfer & 0.852 \interval{0.081} & 2.175 \interval{0.340} & 0.985 \interval{0.074} & 2.282 \interval{0.313} & 0.716 \interval{0.063} & 1.906 \interval{0.282} & 0.913 \interval{0.056} & 2.391 \interval{0.160}  \\ 
 
    \addlinespace[0.3em]
    w/o inter-mesh NNs & 0.444 \interval{0.063} & 0.960 \interval{0.174} & 0.438 \interval{0.052} & 0.958 \interval{0.142} & 0.390 \interval{0.051} & \textbf{0.892} \interval{0.146} & 0.396 \interval{0.049} & 0.946 \interval{0.168}\\ 
 
    \addlinespace[0.3em]
    Ellipsoid template & 0.459 \interval{0.065} & 0.970 \interval{0.145} & 0.452 \interval{0.071} & 0.954 \interval{0.140} & 0.407 \interval{0.044} & 0.948 \interval{0.145} & 0.412 \interval{0.053} & 0.983 \interval{0.201}  \\ 
    
    \addlinespace[0.3em]
    w/o voxel decoder & 0.413 \interval{0.069} & 0.914 \interval{0.168} & 0.424 \interval{0.065} & 0.928 \interval{0.150} & 0.392 \interval{0.057} & 0.916 \interval{0.147} & 0.400 \interval{0.059} & \textbf{0.942} \interval{0.180}  \\ 
 
    \addlinespace[0.3em]
    Lap. on abs. coord. & 0.467 \interval{0.075} & 0.958 \interval{0.150} & 0.444 \interval{0.065} & 0.952 \interval{0.140} & 0.414 \interval{0.050} & 1.102 \interval{0.182} & 0.425 \interval{0.050} & 1.057 \interval{0.178}  \\ 
    
    % Line added after rebuttal
    \addlinespace[0.3em]
    MLP deform  & 0.538 \interval{0.062} & 1.237 \interval{0.195} & 0.542 \interval{0.057} & 1.228 \interval{0.181} & 0.533 \interval{0.057} & 1.472 \interval{0.227} & 0.566 \interval{0.055} & 1.447 \interval{0.218} \\
    \bottomrule
\end{tabular}
\caption{Ablation study results in terms of \acrfull{assd} and \acrfull{hd}. An asterisk (*) indicates that a template with fewer vertices (${\approx}42,000$ instead of ${\approx}168,000$) has been used during testing. Best values are \textbf{highlighted}. See \Cref{sec:ablation} for a detailed description of the individual variations. All values are in mm.}
\label{tab:ablation study}
\end{table*}

\noindent
\textbf{ADNI}
The Alzheimer's Disease Neuroimaging Initiative (ADNI) (\url{http://adni.loni.usc.edu}) provides MRI T1 scans for subjects with Alzheimer's Disease, Mild Cognitive Impairment, and healthy subjects.
After removing data with processing artifacts, we split the data into training, validation, and testing set, balanced according to diagnosis, age, and sex. As ADNI is a longitudinal study, we only use the initial (baseline) scan for each subject. 
In our experiments, we use two different splits of the ADNI data. ADNI$_\text{small}$ contains 299 subjects for training and 60 for validation and testing, respectively. We use it to conduct the experiments for our architecture ablation study. 
A second, larger split ADNI$_\text{large}$ contains 1,155 subjects for training, 169 for validation, and 323 for testing.  ADNI$_\text{large}$ will be used to compare our model with state-of-the-art methods.

\noindent
\textbf{OASIS}
The OASIS-1 dataset~\cite{oasis} contains MRI T1 scans of 416 subjects. 100 subjects have been diagnosed with very mild to moderate Alzheimer's disease. We split the data balanced on diagnosis, age, and sex, resulting in 292, 44, and 80 subjects for training, validation, and testing. We used a small subset of OASIS, the MALC dataset~\cite{landman2012}, for hyperparameter tuning as described in the supplement. 

\noindent
\textbf{Test-retest}
The \gls{trt}~\cite{maclaren2014} contains 120 MRI T1 scans from 3 subjects, where each subject has been scanned twice in 20 days.

\subsection{Ablation Study} \label{sec:ablation}
We evaluate the individual design choices in Vox2Cortex and focus on characteristics of  explicit image-based surface reconstruction methods. 
Especially the combination of a \gls{cnn} and a \gls{gnn} poses the question of how to pass information in the form of features from one subnetwork to the other (in our case from the \gls{cnn} to the \gls{gnn}). For an encoder-decoder \gls{cnn}, for example, it is possible to use extracted feature maps from the encoder~\cite{kong2021b}, from the decoder~\cite{wickramasinghe2020}, or from both. The impact of this and several other choices is hard to assess from existing literature and has not yet been studied extensively. We address this question in our ablation study on the ADNI$_\text{small}$ subset. 
More precisely, we compare Vox2Cortex with the following modifications and present quantitative measures in terms of \gls{assd}, called $D_{23}$ in~\cite{dubuisson1994}, and \gls{hd}~\cite{huttenlocher1993} in \Cref{tab:ablation study}.

\textbf{Voxel2Mesh}: We replace our network architecture with the Voxel2Mesh network~\cite{wickramasinghe2020}. At each mesh deformation step, we sample the \gls{cnn} features from the corresponding voxel-decoder stage. We double the number of voxel-decoder channels at each stage compared to our model for a fair comparison since those are the only ones passed to the \gls{gnn} in Voxel2Mesh. Here, we also use \gls{lns} instead of trilinear interpolation at vertex locations as in~\cite{wickramasinghe2020}.
\textbf{Encoder features}: We sample \gls{cnn} features at each mesh-deformation stage from the corresponding voxel-encoder stage only, see for example~\cite{kong2021b,kong2021}.
\textbf{Classic Chamfer}: We train our architecture with the classic Chamfer loss, which is equal to setting $\kappa(\cdot) \equiv 1$ in \Cref{eq:weighted_Chamfer distance}.
\textbf{W/o inter-mesh NNs}: The exchange of nearest-neighbor vertex positions between white and pial surfaces is omitted.
\textbf{Ellipsoidal template}: Start the deformation from an ellipsoidal template instead of our smoothed cortex template.
\textbf{W/o voxel decoder:} Omit the voxel decoder entirely and sample \gls{cnn} features from the encoder at the respective stage, similar to~\cite{wang2018}.
\textbf{Lap. on abs. coord.}: Compute the Laplacian loss on absolute vertex coordinates instead of relative displacements, i.e., smooth the mesh instead of the displacement field.
% Line added after rebuttal
\textbf{MLP deform}: Every layer of the GNN replaced with a linear layer.

\begin{table*}
\footnotesize
\setlength{\tabcolsep}{4.6pt} % Reduce white-space between columns, SHOULD BE SET AFTER TABLE IS FINISHED

%0.453 \interval{0.072} 
\begin{tabular}{llcccccccc}
    \toprule
    & & \multicolumn{2}{c}{Left WM Surface} 
    & \multicolumn{2}{c}{Right WM Surface}
    & \multicolumn{2}{c}{Left Pial Surface}
    & \multicolumn{2}{c}{Right Pial Surface}

    \\
    \cmidrule(lr){3-4} \cmidrule(lr){5-6} \cmidrule(lr){7-8} \cmidrule(lr){9-10} 
    Data & Method & \gls{assd} & \gls{hd}  & \gls{assd} & \gls{hd}  & \gls{assd} & \gls{hd} & \gls{assd} & \gls{hd}\\\midrule
    
%     \multirow{3}{*}{\makecell{ADNI\\orig}} & \gls{exprecs} & 0.359 \interval{0.116} & 0.795 \interval{0.332} & 0.356 \interval{0.107} & 0.777 \interval{0.317} & 0.314 \interval{0.084} & 0.711 \interval{0.236} & 0.311 \interval{0.079} & 0.699 \interval{0.225} 
%     %& 0.335 \interval{0.100} & 0.745 \interval{0.285} \\ 
%   \\
%     \addlinespace[0.3em]
%     & DeepCSR~\cite{santacruz2021}*  & 0.267 \interval{0.216} & 0.562 \interval{0.725} & 0.260 \interval{0.162} &  0.542 \interval{0.523}& 0.298 \interval{0.184}  & 0.654 \interval{0.596} & 0.294 \interval{0.155}& 0.651 \interval{0.530} \\
%     \addlinespace[0.3em]
%     & nnUNet &1.146 \interval{0.203} & 1.674 \interval{1.481}  & 1.162 \interval{0.317}  & 1.865 \interval{2.808}  & 1.120 \interval{0.169}  & 2.391 \interval{1.267} & 1.191 \interval{0.129}& 2.418 \interval{0.951} \\
%     \addlinespace[0.3em]
    
    %\midrule
    \multirow{3}{*}{\makecell{ADNI\\large}} & Vox2Cortex & \textbf{0.345} \interval{0.056} & \textbf{0.720} \interval{0.125} & \textbf{0.347} \interval{0.046} & \textbf{0.720} \interval{0.087} & \textbf{0.327} \interval{0.031} & \textbf{0.755} \interval{0.102} & \textbf{0.318} \interval{0.029} & \textbf{0.781} \interval{0.102} 
    %& 0.334 \interval{0.044} & 0.744 \interval{0.108} \\ 
   \\
    \addlinespace[0.3em]
    & DeepCSR~\cite{santacruz2021}  & 0.422 \interval{0.058} &0.852 \interval{0.134} & 0.420 \interval{0.058}  & 0.880 \interval{0.156} & 0.454 \interval{0.059} & 0.927 \interval{0.243} & 0.422 \interval{0.053}& 0.890 \interval{0.197}\\
    \addlinespace[0.3em]
    & nnUNet~\cite{isensee2020} & 1.176 \interval{0.345}& 1.801 \interval{2.835} &1.159 \interval{0.242} & 1.739 \interval{1.880}   & 1.310 \interval{0.292}  &  3.152 \interval{2.374}  & 1.317 \interval{0.312}  &3.295 \interval{2.387}    \\
    \addlinespace[0.3em]
    
    \midrule
    \multirow{2}{*}{OASIS} & Vox2Cortex & \textbf{0.315} \interval{0.039} & \textbf{0.680} \interval{0.137} & \textbf{0.318} \interval{0.048} & 0.682 \interval{0.151} & \textbf{0.362} \interval{0.036} & \textbf{0.894} \interval{0.141} & \textbf{0.373} \interval{0.041} & \textbf{0.916} \interval{0.137}\\ 
 
    \addlinespace[0.3em]
    & DeepCSR~\cite{santacruz2021}  & 0.360 \interval{0.042}&0.731 \interval{0.104} & 0.335 \interval{0.050}& \textbf{0.670} \interval{0.195}  &  0.458 \interval{0.056} & 1.044 \interval{0.290}  & 0.442 \interval{0.058}  & 1.037 \interval{0.294}  \\
    \bottomrule
\end{tabular}
\caption{Comparison of Vox2Cortex with DeepCSR and nnUNet on the ADNI$_\text{large}$ and OASIS datasets for the four surfaces. All values are in mm.\vspace{-3mm}}
\label{tab:comparison}
\end{table*}

\Cref{tab:ablation study} shows that the combination of design choices in Vox2Cortex yields the best performance. The curvature-weighted Chamfer loss is the most important ingredient since reconstruction accuracy drops considerably when it is replaced by the standard Chamfer loss. It is worth noticing that whether sampling the image features from the encoder or the decoder has negligible impact on the overall performance. This could be a possible explanation why both approaches have succeeded in the past~\cite{kong2021b,wickramasinghe2020}. Moreover, the voxel decoder can even be omitted entirely without a drastic loss in segmentation accuracy. This might be interesting for practitioners to reduce the memory demand, especially during training.

\subsection{Comparison with Related Work}
We compare Vox2Cortex with DeepCSR~\cite{santacruz2021} on the ADNI$_\text{large}$ and OASIS datasets. For DeepCSR, we sample points with 0.5mm distance to obtain high-resolution predictions.
We chose 3D nnUNet~\cite{isensee2020} for a comparison with a voxel-based segmentation method, as it is a state-of-the-art segmentation model that has been shown to work on multiple segmentation tasks. 
We perform topology correction on the predicted segmentation maps, following the post-processing in~\cite{santacruz2021}, and use marching cubes~\cite{lewiner2003efficient} to extract meshes. 
Table~\ref{tab:comparison} shows that nnUNet achieves less accurate predictions, which is expected as the method is solely voxel-based and the resolution of the generated meshes depends on the image resolution.
Vox2Cortex achieves superior results over DeepCSR in almost all metrics on both datasets. 
In the supplement, we show more results on mesh topology and number of vertices for DeepCSR and Vox2Cortex.

\subsection{Consistency}
We analyze the consistency of Vox2Cortex, DeepCSR (both trained on ADNI$_\text{large}$), and FreeSurfer on the \gls{trt} dataset~\cite{maclaren2014}. To this end, we compute cortical surfaces of brain scans of the same subject from the same day and compare them with respect to their \gls{assd} and \gls{hd}.
Since the morphology of the brain does not change within two consecutive scans of the same day, the resulting reconstructions should be identical up to some variations resulting from the imaging process. We follow~\cite{santacruz2021} in this experiment and align pairs of images with the \gls{icp} before comparing them. 
The results in \Cref{tab:consistency} demonstrate that Vox2Cortex has better reproducibility than DeepCSR and Freesurfer. Further, we compare the inference time of the methods and find that Vox2Cortex is around 25 times faster than DeepCSR.

\begin{table*}
\centering
\begin{tabular}{lccccr}
    \toprule
        Method & \gls{assd} (mm) & \gls{hd} (mm) & $>1$mm & $>2$mm & Inference time \\
    \midrule
        Vox2Cortex (ours) & 0.228 \interval{0.048} & 0.478 \interval{0.101} & 0.80\% & \textbf{0.06}\% & 18.0s  \\
        Vox2Cortex* (ours) & \textbf{0.225} \interval{0.049} & \textbf{0.471} \interval{0.103} & \textbf{0.77}\% & \textbf{0.06}\% & 2.1s \\
        DeepCSR &0.357 \interval{0.284} & 0.739 \interval{0.595}& 5.82\%& 2.23\%& 445.7s\\
        FreeSurfer & 0.291 \interval{0.133} & 0.605 \interval{0.279} & 2.87\% & 0.67\% & \textgreater 4h\\
    \bottomrule
\end{tabular}
\caption{Comparison of reconstruction consistency in terms of \gls{assd} and \gls{hd} on the \gls{trt} dataset. In addition, we provide the inference time of the respective method per 3D scan. An asterisk (*) indicates smaller templates (${\approx}42,000$ instead of ${\approx}168,000$ vertices per surface).}
\label{tab:consistency}
\end{table*}

\subsection{Cortical Thickness}
\begin{figure}
    \centering
    \includegraphics[width=\linewidth]{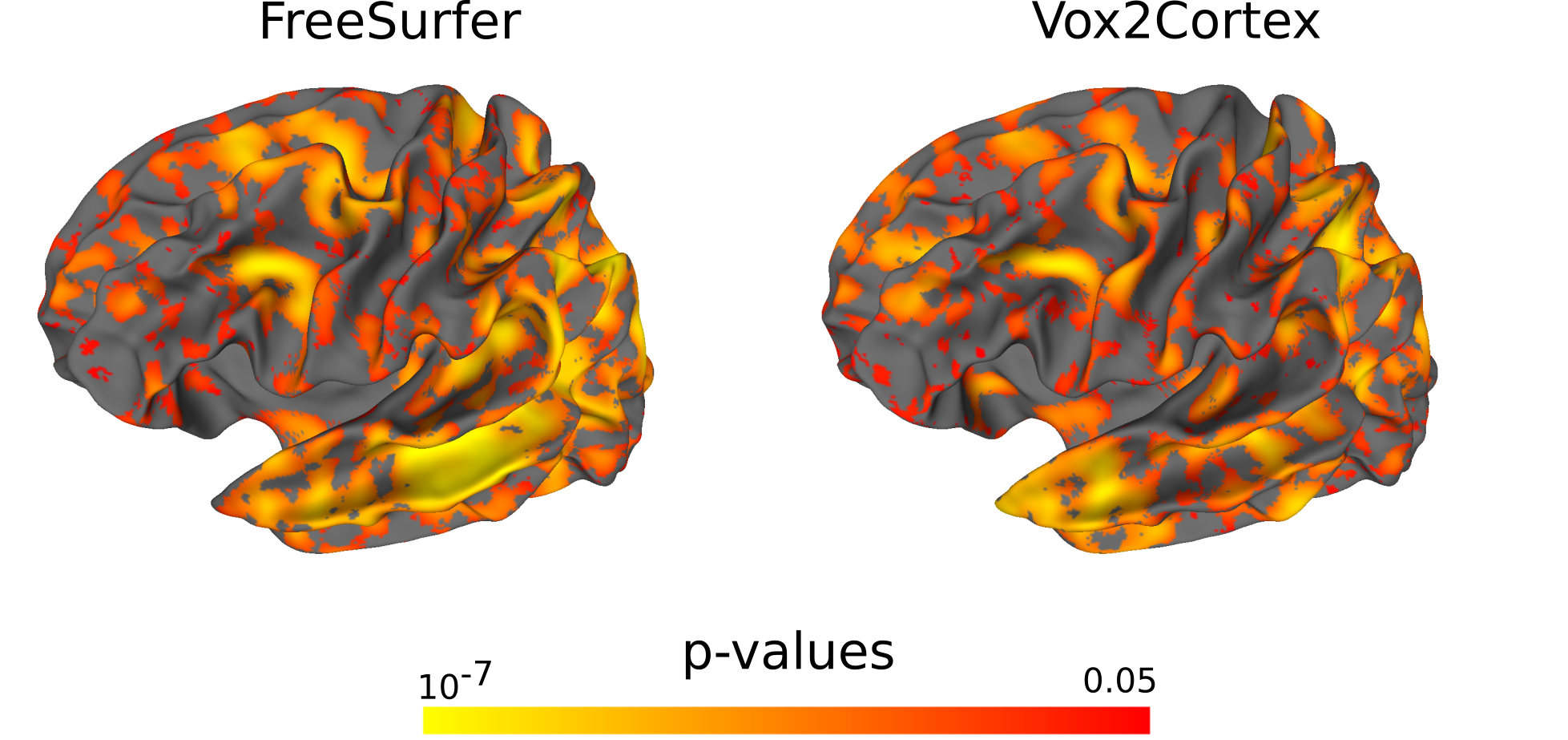}
    \caption{Group comparison of cortical atrophy in the left hemisphere between patients diagnosed with Alzheimer's disease and healthy controls on the  ADNI$_\text{large}$ test-split.\vspace{-3mm}}
    \label{fig:thickness_comparison}
\end{figure}

The surface reconstruction enables the computation of cortical thickness, which is an important biomarker for assessing cortical atrophy in neurodegenerative disorders. In this experiment, we compute cortical thickness measurements from Vox2Cortex surfaces and compare them against FreeSurfer thickness measures.
The thickness is computed by means of the closest-point correspondences between the WM and pial surfaces~\cite{miller2000}. 
The median cortical thickness error per vertex on the OASIS test set compared to closest FreeSurfer vertices is 0.305mm [lower quartile: 0.140mm, upper quartile: 0.564mm]. For comparison, atrophy in regions affected by Alzheimer's disease often lies in the range of one millimeter or more~\cite{lerch2005}. In the supplement, we also visualize the respective measurements on one subject from the OASIS dataset.
On the ADNI$_\text{large}$ test-split, we performed a group analysis by testing individual vertices for significantly lower thickness in subjects diagnosed with Alzheimer's disease ($n=50$) compared to healthy controls ($n=124$). For this purpose, we leveraged the FreeSurfer FsAverage template to have all thickness measurements in the same domain and perform statistical tests per vertex. We also computed cortical thickness on FreeSurfer meshes with closest-point correspondences 
%(pial to WM)
instead of using the default thickness measurements to have a fair comparison. \Cref{fig:thickness_comparison} visualizes the p-values (t-test, one sided) for the left hemisphere. 
As there is a high similarity between significance maps created by Vox2Cortex and FreeSurfer, we conclude that Vox2Cortex is well-suited for group analysis for cortical thickness. 

\section{Limitations and Potential Negative Impact}
All presented results were created with the unmodified output of our model. However, we would like to mention that we cannot guarantee that predicted meshes have no self-intersections. This property might be necessary for some tasks, e.g., volume measurements. However, all self-intersections can be easily removed with algorithms like MeshFix~\cite{attene2010} (runtime app.\ 16s for a surface with 168,000 vertices), keeping the remaining mesh unchanged.
As there exist no manually generated ground-truth surfaces for our data, we used FreeSurfer surfaces as pseudo ground-truth as in~\cite{santacruz2021,ma2021b}. Although we identified and removed some faulty labels from our training set, some labels might still be noisy. 
%\paragraph{Potential negative impact}
Our method has the potential to help radiologists to visualize the brain surfaces and to compute measurements like cortical thickness quickly. The predictions should, however, not be used to make clinical decisions. Our model has only been tested on the data discussed in this work, and we cannot guarantee that it will perform as well on unseen data from different domains. Further, the data analyzed in this study only includes healthy subjects and subjects with dementia. If other morphological changes in the brain are present (e.g., tumors), the trained model might provide wrong predictions.
We have balanced our data splits with respect to patient's age and sex to avoid bias, but our model might still lack fairness and might discriminate against groups of people underrepresented in the given datasets. 

\section{Conclusion}
\label{sec:conclusion}
In this work, we have presented a novel method for the simultaneous reconstruction of white matter and pial surfaces from brain \gls{mr} images. For the first time, we start from a general brain template and deform it in multiple iterations leveraging thoroughly combined convolutional and graph convolutional neural networks. To achieve a high reconstruction accuracy in densely folded regions, we successfully exploit ground-truth curvatures in a novel curvature-weighted Chamfer-loss function. We believe that this loss can also be helpful in other medical and non-medical fields where complex 3D geometries appear. Our experiments show that the proposed combination of loss functions directly yields state-of-the-art cortical surfaces as the output of an end-to-end trainable architecture, while being orders of magnitude faster. 
Finally, we demonstrated that accurate cortical thickness maps can be derived for studying atrophy in Alzheimer's disease. 

\vspace{-3mm}
{\small
\paragraph{Acknowledgment}
This research was partially supported by the Bavarian State Ministry of Science and the Arts and coordinated by the bidt, and the BMBF  (DeepMentia, 031L0200A). We gratefully acknowledge the computational resources provided by the Leibniz Supercomputing Centre (www.lrz.de).
}

%%%%%%%%% REFERENCES
{\small
\bibliographystyle{ieee_fullname}
\bibliography{egbib}
}
\setcounter{equation}{0}
\setcounter{figure}{0}
\setcounter{table}{0}

\newpage

\twocolumn[{%
 \centering

\Large
    \textbf{Vox2Cortex: Fast Explicit Reconstruction of Cortical Surfaces from \\ 3D MRI Scans with Geometric Deep Neural Networks --- Supplementary Material}
     \lineskip .5em
      % additional small space at the end of the author name
      \vskip .5em
      % additional empty line at the end of the title block
      \vspace*{12pt}

}]

\appendix

\section{Proof for Cuvature-Weighted Chamfer}

\begin{figure}
    \centering
    \resizebox{75mm}{!}{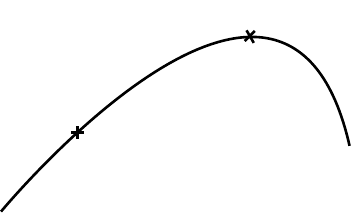}
    \caption{Ground-truth points $a$ and $b$ with curvature $\kappa(a) < \kappa(b)$ and predicted points $u$ and $v$.}
    \label{fig:skizze}
\end{figure}
We want to give a brief mathematical intuition why our curvature-weighted Chamfer loss emphasizes geometric accuracy in high-curvature regions compared to low-curvature regions. Imagine therefore two ground-truth points $a$ and $b$ with respective curvature $\kappa(a) < \kappa(b)$ and closest predicted points $u$ and $v$ as shown in \Cref{fig:skizze}. Furthermore, let the distance from the prediction to the ground truth be equal in both cases, such that $\norm{u-a} = \norm{v-b}$. For the sake of simplicity, we treat the predicted values $u$ and $v$ as the parameters that are optimized by gradient descent, i.e., $u' = u - \lambda \frac{\partial \gls{LCDcurv}(a,u)}{\partial u}$ with learning rate $\lambda > 0$. Based on Equation (7) in the main paper, the gradient of the curvature-weighted Chamfer loss with respect to $u$ calculates as
\begin{align}
    \begin{split}
    \frac{\partial \gls{LCDcurv}(a,u)}{\partial u} 
    &= \frac{\partial}{\partial u} \left[ \kappa(a) \left( \norm{a-u}^2 + \norm{u-a}^2 \right) \right]\\
    &= 4 \kappa(a) (u-a).
    \end{split}
\end{align}
The calculation of $\frac{\partial \gls{LCDcurv}(b,v)}{\partial v} = 4 \kappa(b) (v-b)$ works analogously. The parameter updates are given by
\begin{align}
    \begin{split}
        u' &= u - \frac{\partial \gls{LCDcurv}(a,u)}{\partial u} = u + 4 \lambda \kappa(a) (a-u), \\
        v' &= v - \frac{\partial \gls{LCDcurv}(b,v)}{\partial v} = v + 4 \lambda \kappa(b) (b-v).
    \end{split}
\end{align}
Further, we have $\norm{a-u} = \norm{b-v}$ and $\kappa(a) < \kappa(b)$, and thus we get $\norm{v'-b} < \norm{u'-a}$ if we assume that we don't ``shoot over'' the goal, i.e., $0 < 4 \lambda \kappa(a) < 4 \lambda \kappa(b) < 1$. That is, point $v$ is pushed more towards $b$ compared to $u$ towards $a$ within one backward pass. \hfill $\blacksquare$

% stuff for supplementary material
\section{Definitions of Loss Functions}
\paragraph{Binary cross entropy} The cross-entropy loss between a predicted binary segmentation $\pred{B}_l \in [0, 1]^{\gls{H_img}  \gls{W_img}  \gls{D_img}}$ and a label $\gt{B} \in \{0, 1\}^{\gls{H_img}  \gls{W_img}  \gls{D_img}}$, where voxels are enumerated from 1 to $N=HWD$, is defined as
\begin{align} \label{eq:cross entropy}
    \begin{split}
	\gls{LCE}(\pred{B}_l, \gt{B}) &= - \frac{1}{\gls{N_img}} \sum_{i = 1}^{\gls{N_img}}
	\left[\gt{B}(i) \log \pred{B}_l(i)\right.\\ 
	&+{} 
	\left.(1 - \gt{B}(i))(1-\log \pred{B}_l(i))\right],
	\end{split}
\end{align}
where $B(i)$ is the value of voxel $i$.

\paragraph{Inter-mesh normal consistency loss}
While the Chamfer distance takes into account the spatial position of two meshes, i.e., enforcing surface points to lie ``at the right location'', the \emph{cosine distance} considers the orientation of meshes. In general, one can compute the cosine distance within one mesh, which we refer to as \emph{intra-mesh normal consistency}, and between two meshes, which we call \emph{inter-mesh normal consistency}.

The inter-mesh normal consistency loss is defined based on the normal vectors of adjacent points in the predicted and the ground-truth mesh. Let $\Ppred_{s,c}, \Pgt_c$ be predicted and ground-truth point sets with associated normals $\pred{\mathcal{N}_{s,c}}=\{\vec{n}(\vec{p}) | \vec{p} \in \Ppred_{s,c} \}$ and $\gt{\mathcal{N}_c} = \{\vec{n}(\vec{p}) | \vec{p} \in \Pgt_c \}$, respectively. Then, the inter-mesh normal consistency loss is given by
\begin{align} \label{eq:normal consistency}
	\begin{split}
	\gls{Lcos} (\Mpred_{s,c}, \Mgt_c) 
	&= \frac{1}{\abs{\Pgt_c}} \sum_{\vec{u} \in \Pgt_c} 1 - \cos(\vec{n}(\vec{u}), \vec{n}(\tilde{\vec{v}})) \\
	&+ \frac{1}{\abs{\Ppred_{s,c}}} \sum_{\vec{v} \in \Ppred_c} 1 - \cos(\vec{n}(\vec{v}), \vec{n}(\tilde{\vec{u}})),
\end{split}
\end{align}
where $\tilde{\vec{v}} = \underset{\vec{r} \in \Ppred_{s,c}}{\argmin} \norm{\vec{u} - \vec{r}}^2$ and $\tilde{\vec{u}} = \underset{\vec{r} \in \Pgt_c}{\argmin} \norm{\vec{v} - \vec{r}}^2$.

In other words, each normal vector at a certain point $\vec{p}$ is compared to the normal belonging to the nearest neighbor of $\vec{p}$ in the respective other point set. Since nearest-neighbor correspondences are also required for the computation of the Chamfer loss, we use the same point sets $\Ppred$, $\Pgt$ for the computation of the Chamfer and inter-mesh normal consistency loss in practice (see paragraph ``Curvature-weighted Chamfer loss'' in the main paper for details about the point sets).

\paragraph{Intra-mesh normal consistency loss}
Instead of computing the cosine distances among the normals of two different meshes as described above, it is also possible to compare normal vectors of two adjacent faces of the same mesh. Two faces $f_1$ and $f_2$ are adjacent if they share a common edge $e = f_1 \cap f_2$. This intra-mesh normal-consistency, which we denote as $\gls{LNC}$, intuitively measures the smoothness of a mesh as it is lowest for meshes with no curvature. Formally, it is defined as
\begin{equation}
	\gls{LNC}(\Mpred_{s,c}) = \sum_{\substack{e = f_1 \cap f_2 \neq \emptyset  \\ f_1, f_2 \in \pred{\gls{F}}_{s,c}}} 1 - \cos(\vec{n}(f_1), \vec{n}(f_2)),
\end{equation}
where $\vec{n}(f)$ assigns a normal to each face of the mesh. As this loss is only computed based on a predicted mesh not taking into account any ground truth, it belongs to the group of mesh-regularization losses.

\paragraph{Laplacian loss}
Another measure for the smoothness of a mesh is computed based on the uniform Laplacian operator $\gls{Lap} = \gls{D}^{-1} \gls{A} - \mat{I}$, where \gls{D} is the degree and \gls{A} is the adjacency matrix of the mesh. More precisely, \emph{Laplacian smoothing} is defined as
\begin{equation} \label{eq:rel_laplacian loss}
	\gls{LLaprel}(\Mpred_{s,c}) = \frac{1}{\abs{\pred{\gls{V}}_{s,c}}} \sum_{i=1}^{\abs{\pred{\gls{V}}_{s,c}}} \norm{(\pred{\gls{Lap}}_{s,c}. \pred{\gls{Vrel}}_{s,c})_i}
\end{equation}
This is a well-known objective for smooth meshes~\cite{nealen2006}. While many works~\cite{wang2018, wickramasinghe2020, kong2021} smooth the mesh with respect to vertex coordinates $\pred{\gls{V}}_{s,c}$, we got inspired by~\cite{zhao2020} and apply the Laplacian operator to the displacement field $\pred{\gls{Vrel}}_{s,c}$. Our ablation study confirms that this is a good choice. More precisely, $\pred{\gls{Vrel}}_{s,c}$ represents the displacement vectors moving the vertices $\pred{\gls{Vmat}}_{s-1, c}$ to $\pred{\gls{Vmat}}_{s,c}$, i.e., $\pred{\gls{Vmat}}_{s,c} = \pred{\gls{Vmat}}_{s-1, c} + \pred{\gls{Vrel}}_{s, c}$ transforms the mesh $\Mpred_{s-1, c}$ into $\Mpred_{s,c}$.

Even though a Laplacian loss does not guarantee that the predicted meshes are free of self-intersections, 
%(neither in its relative nor in its absolute form)
it generally enforces the predicted meshes to have a smooth surface, i.e., few self-intersections. Also note that in \cref{eq:rel_laplacian loss} $\pred{\gls{Lap}}_{s,c}$ is considered to be a constant, i.e., the loss is not backpropagated through the creation of $\pred{\gls{Lap}}_{s,c}$. 

\paragraph{Edge loss}
Yet another mesh loss function with regularizing purposes is given by the \emph{edge loss}. The edge loss with respect to a predicted mesh is defined as
\begin{equation}
	\gls{Ledge}(\Mpred_{s,c}) = \frac{1}{\abs{\pred{\gls{E}}_{s,c}}} \sum_{(i,j) \in \pred{\gls{E}}_{s,c}} \norm{\vec{v}_i - \vec{v}_j}^2.
\end{equation}
Intuitively, this loss function enforces meshes with homogeneous edge-lengths, leading to a homogeneous distribution of vertices on the surface. In general, this is desirable in the context of cortical surfaces since the folds of the cortex are also distributed homogeneously.

\paragraph{Mesh-loss weights}
We condition the mesh-loss weights on the surface class, even though this increases the number of hyperparameters, as we have found that different weights are necessary for white matter and pial surfaces in order to achieve optimal reconstruction quality. In practice, we tuned the mesh-loss weights for white matter and pial surfaces independently of each other (ignoring the respective other surfaces in those runs and considering only one hemisphere) on the small MALC dataset~\cite{landman2012}. It contains 15 training scans, 7 validation scans, and 8 test scans (which we ignore since testing our model on such a few scans is probably not meaningful). From the tuning, we got the following loss weights:

\begin{minipage}[h!]{\linewidth}
\vspace{12pt}
\begin{tabular}{lccccc}
    \toprule
     Surface & $\gls{mlw}_{1,c}$ & $\gls{mlw}_{2,c}$ & $\gls{mlw}_{3,c}$ & $\gls{mlw}_{4,c}$ & $\gls{mlw}_{5,c}$  \\
     \midrule
     $c=\text{wm}$ & 1.0 & 0.01 & 0.1  & 0.001  & 5.0\\
     $c=\text{pial}$ & 1.0 & 0.0125  & 0.25  & 0.00225 & 5.0\\
     \bottomrule
     \vspace{3mm}
\end{tabular}
\end{minipage}

Mesh-loss function weights for inter-mesh normal consistency \gls{Lcos}, intra-mesh normal consistency \gls{LNC}, and Laplacian smoothing \gls{LLap} were first tuned with a grid search containing 0.1, 0.01, 0.001 and then fine-tuned with the values $x+0.5x$, $x$, $x-0.5x$, where $x$ was the respective best value of the first tuning. Weights for Chamfer and edge losses were set to 1 in this procedure and the edge-loss weight was later tuned separately trying the values 1, 5, and 10.

\section{Implementation Details}
We implemented our method based on pytorch v1.7.1 \url{https://pytorch.org/} and pytorch3d v0.4.0 \url{https://pytorch3d.readthedocs.io}. We ran experiments on NVIDIA Quadro and Titan RTX GPUs with 24GB memory each (one GPU per training). In addition, we used CUDA v10.2.89, CUDNN v7.6.5, python v3.8.8, and the repositories from DeepCSR~\cite{santacruz2021} \url{https://bitbucket.csiro.au/projects/CRCPMAX/repos/deepcsr/browse} and Voxel2Mesh~\cite{wickramasinghe2020} \url{https://github.com/cvlab-epfl/voxel2mesh/blob/master/README.md}.

\section{Hyperparameters}
A list of hyperparameters is in \Cref{tab:hyperparams}. We trained our models for 100 epochs (OASIS and ADNI$_\text{small}$) and 40 epochs (ADNI$_\text{large}$) and chose the best model with respect to the respective validation set in terms of voxel IoU and Hausdorff distance. 

\begin{table*}
\centering
\begin{tabular}{lcccccc}
    \toprule
        Optimizer & 
        \makecell{\gls{cnn} learn-\\ing rate} &
        \makecell{\gls{gnn} learn-\\ing rate} &
        Batch size & 
        \makecell{Mixed\\ precision} & 
        CNN channels & 
        GNN channels \\
        \midrule
        \makecell{Adam~\cite{kingma2015}\\ ($\beta_1 = 0.9$,\\ $\beta_2 = 0.999$)} &
        0.0001 & 
        0.00005 & 
        \makecell{2\\ (1 for OASIS)} & 
        yes &
        \makecell{16, 32, 64, 128 , 256,\\ 64, 32, 16, 8} & 
        255, 64, 64, 64, 64 \\
    \bottomrule
\end{tabular}
\caption{Hyperparameters used in our experiments.}
\label{tab:hyperparams}
\end{table*}

\section{Additional Analysis of Experiments}
\begin{table*}
\centering
\begin{tabular}{lcccccccc}
    \toprule
    & \multicolumn{4}{c}{Pial Surfaces} 
    & \multicolumn{4}{c}{WM Surfaces}\\
    \cmidrule(lr){2-5} \cmidrule(lr){6-9}
        Method & CC & genus & \# faces & \# vertices & CC & genus & \# faces & \# vertices\\
        \midrule
        Ours & 1 & 0 & 336112 & 168058 & 1 & 0 & 336112 & 168058 \\
        DeepCSR & 48.6 & 152.4 & 1341838.3 & 670711.5 & 18.3 & 15.8 & 1209313.5 & 604661.7\\
        DeepCSR + top & 1 & 0 & 1291385.5 & 645694.8 & 1 & 0 & 1160980.8 & 580492.4\\

    \bottomrule
\end{tabular}
\caption{Comparison of topological measures (number of connected components (CC) and genus) and quantification of mesh complexity in number of faces and vertices. We present the average values for white and pial surfaces. We compare predictions of our method to DeepCSR with and without topology correction on the OASIS test-set. For our method the number of faces and vertices is defined by the initial template and does not change.}
\label{tab:topology}
\end{table*}

\begin{figure}
    \centering
    \includegraphics[width=\linewidth]{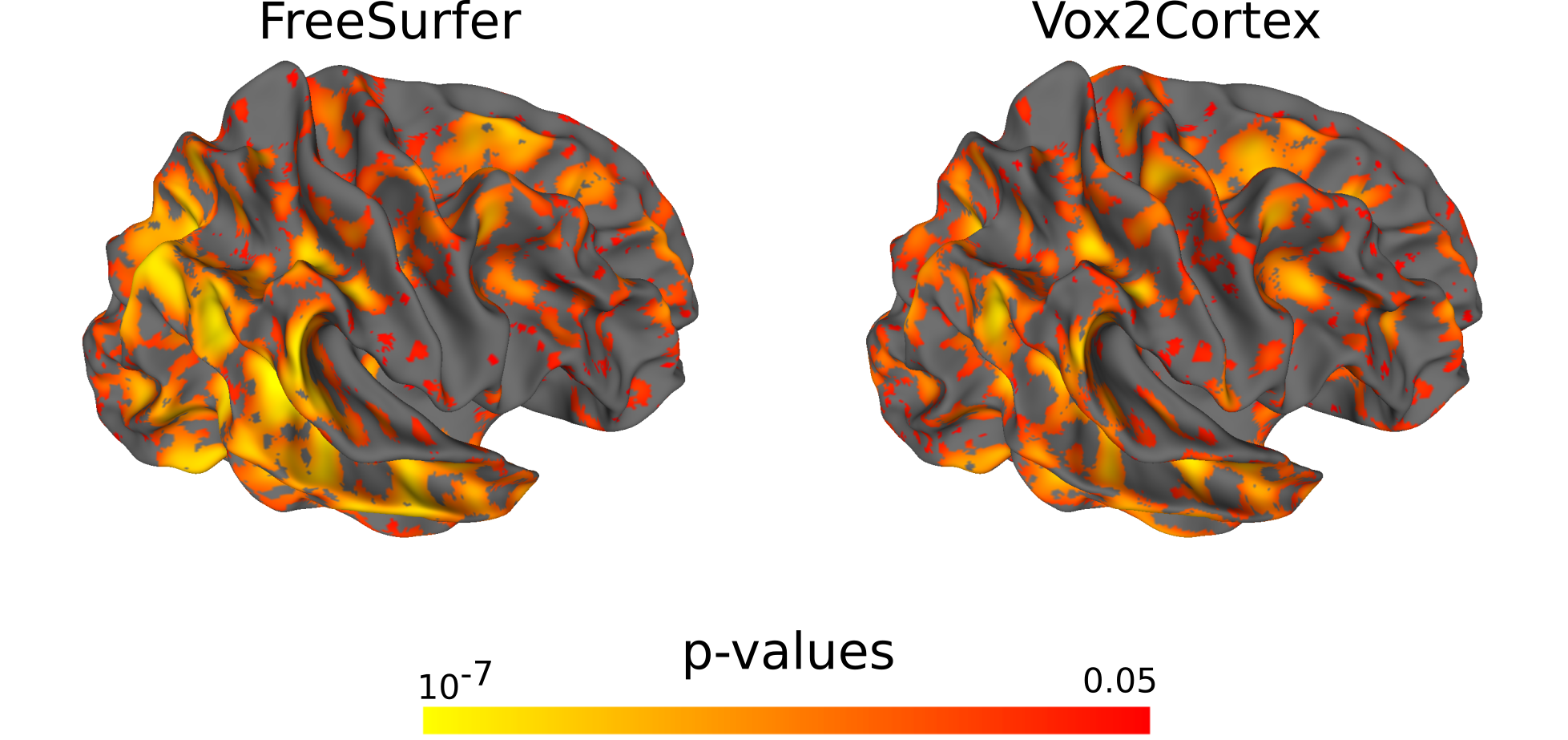}
    \caption{Group comparison of cortical atrophy in the right hemisphere between patients diagnosed with Alzheimer's disease and healthy controls on the  ADNI$_\text{large}$ test-split.}
    \label{fig:rh_thickness_comparison}
\end{figure}

\paragraph{Cortical atrophy}
We show the study of cortical atrophy (Figure 5 in the main paper, left hemisphere) for the right hemisphere in \Cref{fig:rh_thickness_comparison}.

\begin{figure*}[t]
    \centering
    \includegraphics[width=\linewidth]{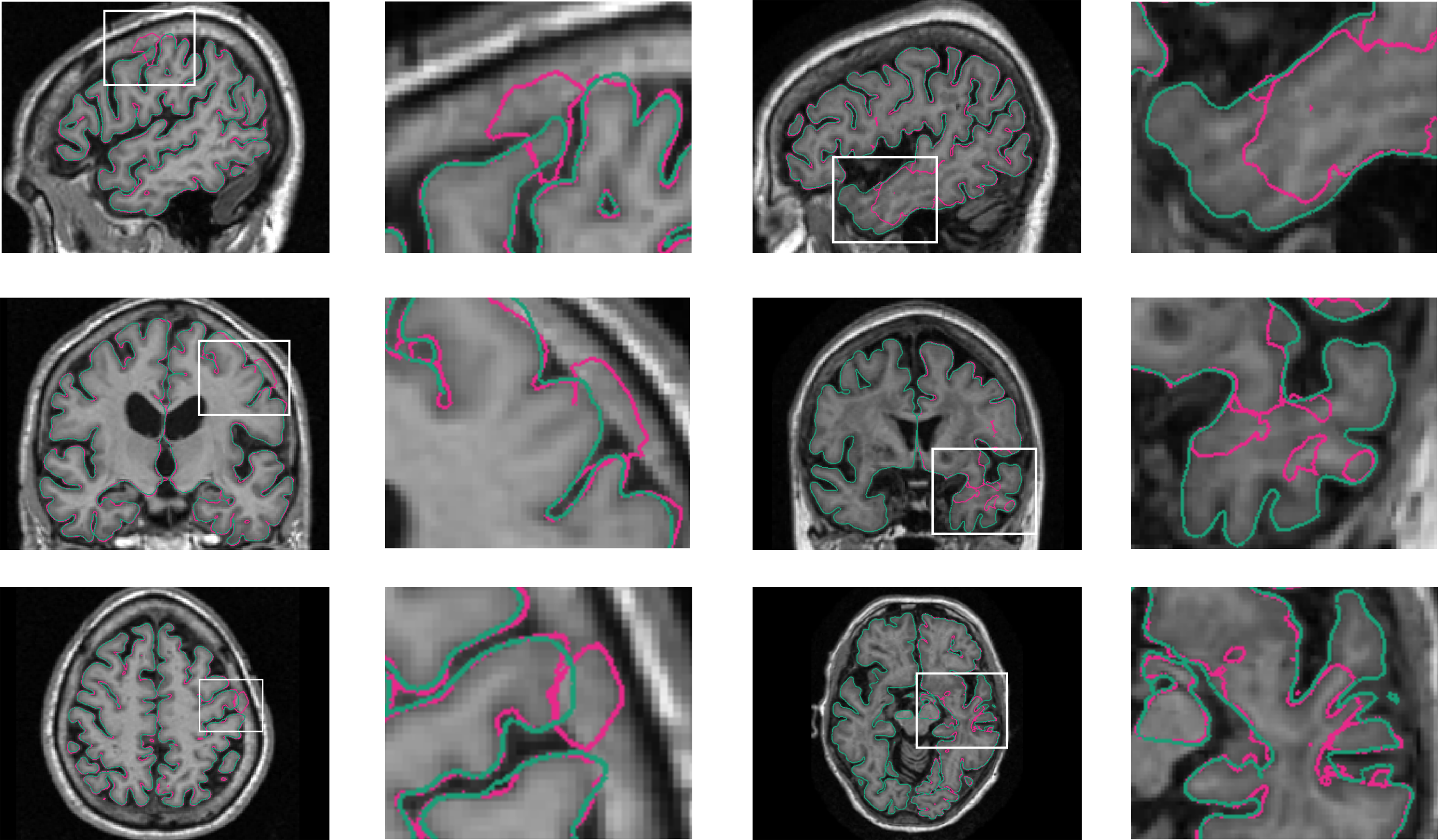}
    \caption{MRI scans with overlaying pial surfaces generated by FreeSurfer (pink) and Vox2Cortex (green). From top to bottom we show sagittal, coronal, and axial slices of two subjects with zoomed in parts where FreeSurfer failed.}
    \label{fig:failed}
\end{figure*}

\paragraph{Visual analysis of Freesurfer fails}
In our ADNI$_\text{large}$ dataset, we removed samples in which FreeSurfer failed. As it is quite difficult to perform automated quality control of the FreeSurfer surface pipeline, we removed all scans that failed in the segmentation of one or more regions as identified by UCSF quality control guidelines~\cite{hartig2014}. 
We then applied the trained model to the previously removed cases where FreeSurfer failed and visualize results in Figure~\ref{fig:failed}, where we focus on pial surfaces due to better visibility. The first case is a mild case where FreeSurfer was able to generate 4 surfaces, but we can observe that the left pial surface extends into the dura. Our model does not produce those artifacts. We further display a more extreme case, where FreeSurfer was not able to generate surfaces for the right hemisphere and also failed to segment parts of the left temporal lobe correctly. 

\begin{figure*}[t]
    \centering
    \includegraphics[width=\linewidth]{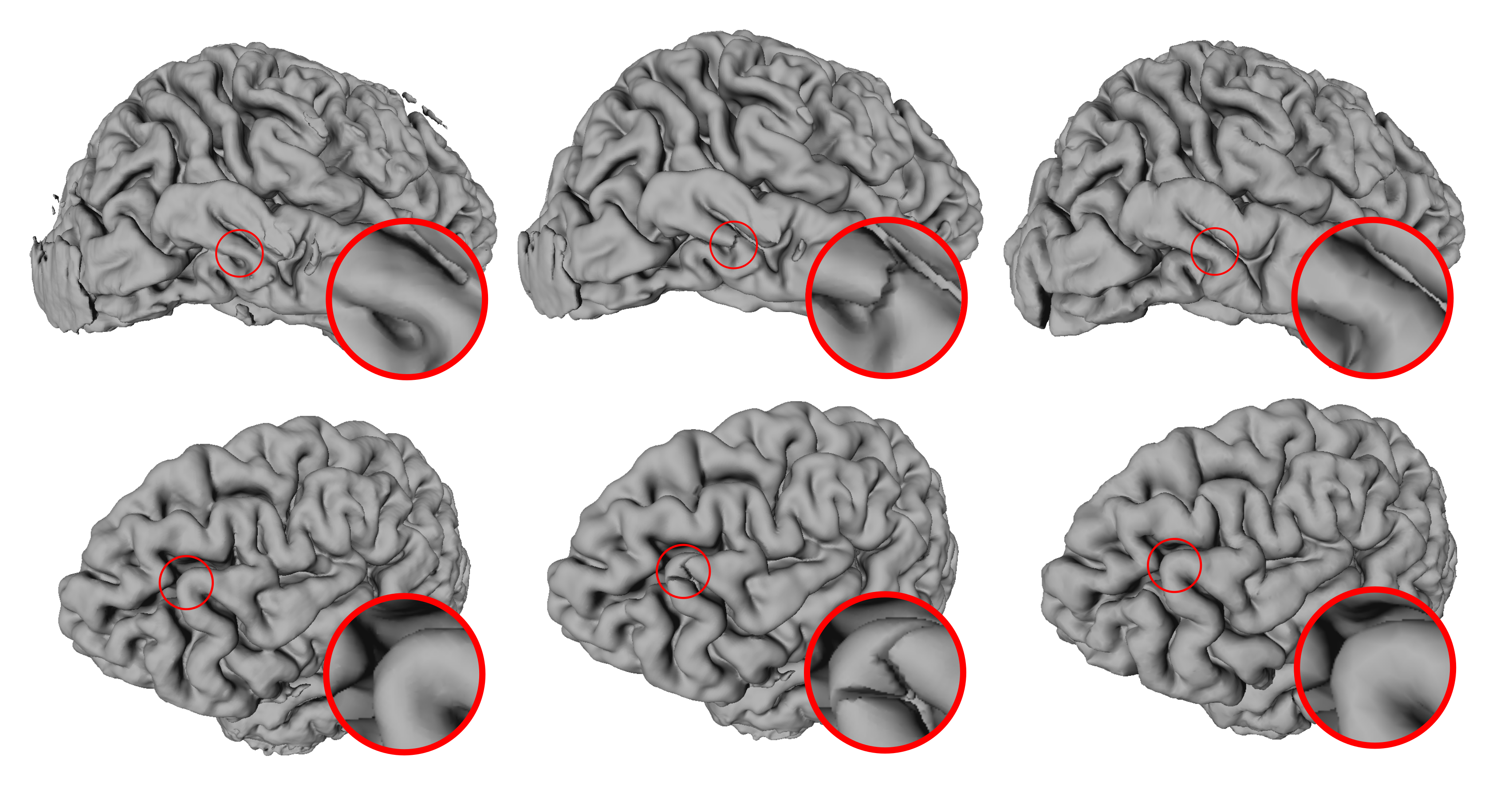}
    \caption{Visualization of incorrect anatomy due to topology correction. We show pial surfaces from 2 different patients from the OASIS dataset. Left: prediction by DeepCSR before topology correction, middle: after topology correction, right: FreeSurfer pseudo ground truth.  }
    \label{fig:top_fail}
\end{figure*}

\begin{figure*}[htb]
    \centering
    % \resizebox{0.75\textwidth}{!}{
    % \input{figures/thickness_figure.pdf_tex}
    % }
    \includegraphics[width=\textwidth]{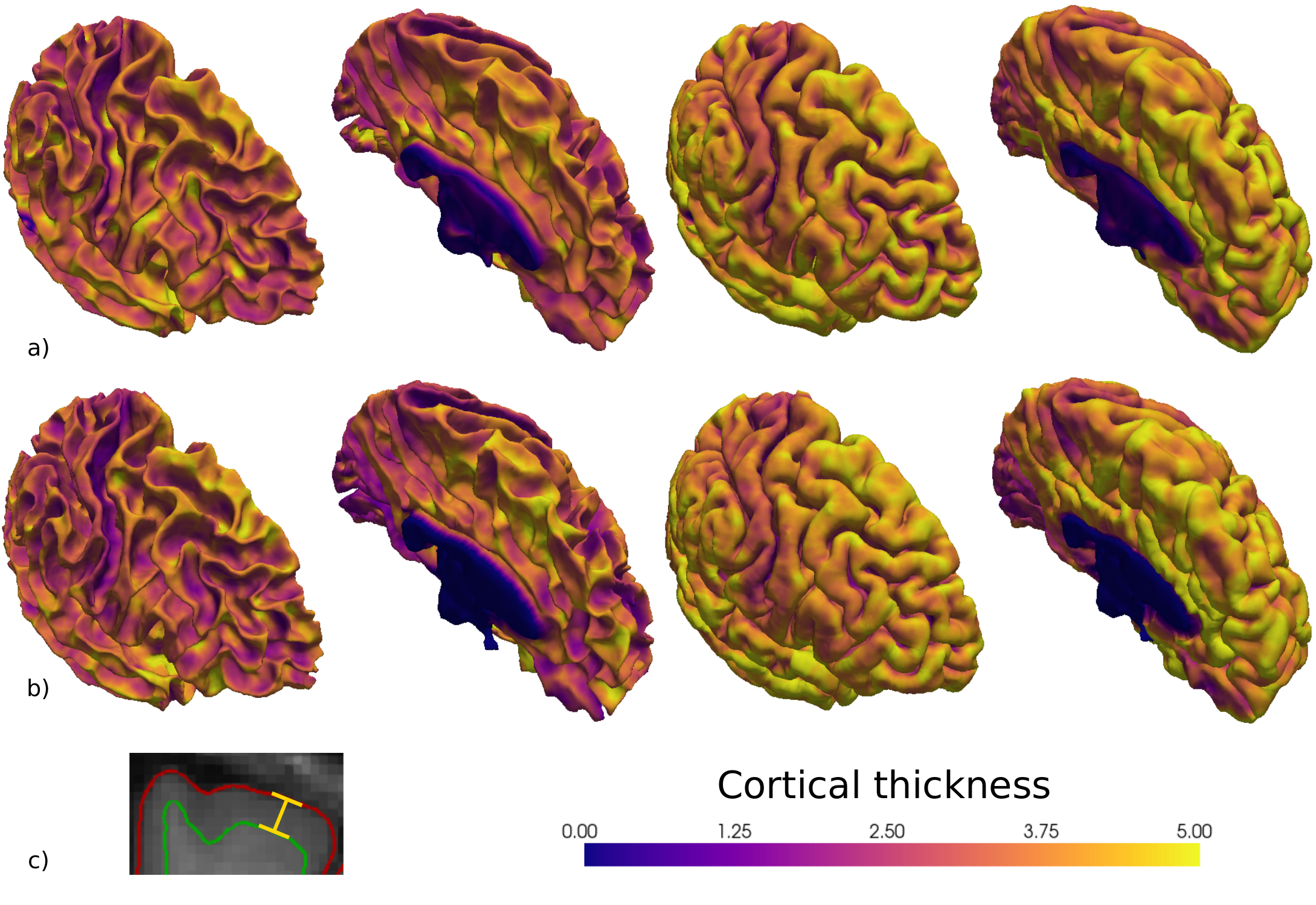}
    \caption{OASIS meshes color-coded with cortical thickness per vertex in mm. a) Vox2Cortex meshes, b) FreeSurfer meshes, c) cortical thickness between white matter (green) and pial (red) surface.}
    \label{fig:thickness_OASIS}
\end{figure*}

\paragraph{Cortical thickness on OASIS}
We visualize thickness measurements on an exemplary subject from the OASIS dataset in \Cref{fig:thickness_OASIS}. It can be well observed that measurements on Vox2Cortex meshes largely coincide with measurements on FreeSurfer pseudo-ground-truth meshes.

\end{document}